\documentclass[10pt,twocolumn,letterpaper]{article}

\usepackage[pagenumbers]{cvpr} 

\usepackage{graphicx}
\usepackage{booktabs}
\usepackage{amsmath}
\usepackage{algorithm}
\usepackage{algorithmic}
\usepackage[accsupp]{axessibility}
\usepackage{makecell}
\usepackage{wrapfig} 
\usepackage{pifont}
\usepackage{multirow}
\usepackage{colortbl}
\usepackage{tikz}
\usepackage{xcolor}
\usepackage{caption} 

\definecolor{gcol}{RGB}{70,0,120}
\definecolor{ccol}{RGB}{90,0,150}
\definecolor{scol}{RGB}{110,0,180}
\definecolor{ecol}{RGB}{130,20,200}
\definecolor{dcol}{RGB}{150,40,215}
\definecolor{icol}{RGB}{170,70,225}
\definecolor{tcol}{RGB}{190,100,235}

\definecolor{cvprblue}{rgb}{0.21,0.49,0.74}
\usepackage[pagebackref,breaklinks,colorlinks,allcolors=cvprblue]{hyperref}

\def\paperID{7570} 
\def\confName{CVPR}
\def\confYear{2026}

\title{
\textcolor{gcol}{G}%
\textcolor{ccol}{e}%
\textcolor{scol}{o}%
\textcolor{ecol}{E}%
\textcolor{dcol}{d}%
\textcolor{icol}{i}%
\textcolor{tcol}{t}:
Geometry-Aware Object Editing via
Dual-Branch Denoising
}

\author{
Yi He\textsuperscript{1,*} \quad
Jiangming Wang\textsuperscript{3,*} \quad
Xinyu Wang\textsuperscript{1} \quad
Mark Fong\textsuperscript{4}  \quad
Songchun Zhang\textsuperscript{5} \quad \\
Yuxuan Xue$^{6,\ddagger}$ \quad 
Hai-Tao Zheng\textsuperscript{1,2,\dag} \quad
Yue Ma\textsuperscript{5,\dag} \\
\textsuperscript{1}Shenzhen International Graduate School, Tsinghua University \quad
\textsuperscript{2}Pengcheng Laboratory \\
\textsuperscript{3}Sun Yat-sen University \quad
\textsuperscript{4}Peking University \quad
\textsuperscript{5}HKUST \quad
\textsuperscript{6}University of Tübingen
}
\begin{document}

\twocolumn[{%
\renewcommand\twocolumn[1][]{#1}%
\maketitle

\begin{center}
    \vspace{-2em}
    \includegraphics[width=0.95\linewidth]{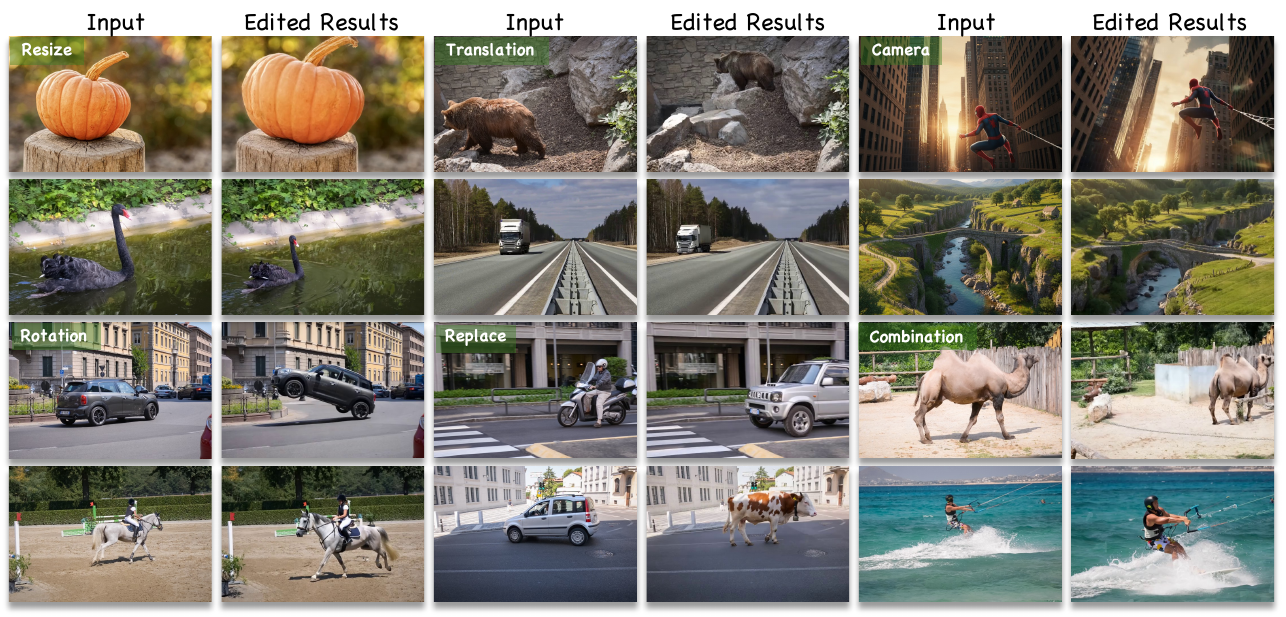}
    \captionof{figure}{\textbf{Showcase of proposed GeoEdit}. In this paper, we propose GeoEdit, a training-free pipeline that lifts editing into 3D for physically plausible object manipulation, without external 3D software or synthetic training data.}
    \label{fig:teaser}
    \vspace{1.5em}
\end{center}%
}]
\begingroup
\renewcommand{\thefootnote}{}
\footnotetext{* Contributed equally.}
\footnotetext{$^{\ddagger}$Project leader.}
\footnotetext{$\dagger$ Corresponding authors.}
\endgroup

\begin{abstract}
Precisely manipulating objects in a single photograph (translation, rotation, scaling) while obeying 3D physical constraints remains unsolved for diffusion-based editors. Current 2D methods lack spatial awareness and produce perspective violations. Forcing structural proxies into the latent space also disrupts variance homogeneity, and the resulting self-attention leakage leads to ghosting and background blur. The core difficulty is asymmetric: the relocated object must follow a rigid geometry, yet the uncovered background needs freedom to synthesize plausible content.

We present \textbf{GeoEdit}, a training-free \emph{Lift-Manipulate-Render-Denoise} pipeline that satisfies both constraints. We decouple scene and object in 3D, align them through point correspondence, and render a geometry-aligned proxy with a structural depth map. A Dual-Branch Denoising stage then refines this proxy: a video diffusion backbone preserves object identity, while 3D constraints are injected into the foreground within a narrow denoising window at matching noise variance (variance-homogeneous injection). The background denoises freely. Because the injected signal matches the native latent statistics, self-attention stays undisturbed. We also introduce GeoEditBench, a pose-aware benchmark covering object translation, object rotation, and camera movement with pose-aware evaluation metrics. Experiments confirm consistent gains in geometric accuracy, identity fidelity, and background quality, validated by automatic metrics and human studies. Our codes are available at \url{https://github.com/Heey731/GeoEdit}.
\end{abstract}

\section{Introduction}
\label{sec01:intro}



Image diffusion models~\cite{ho2020denoising,rombach2022high,song2026vista,song2026streamingeffect,gao2026pai,song2024processpainter, li2025lodge, li2024lodge, li2023finedance} have advanced synthesis and editing, but one task remains open: precisely manipulating objects in 3D, translating, rotating, or scaling them within a photograph while respecting physical constraints. Such control matters for content creation and augmented reality, where edits must obey perspective, occlusion, and scene geometry.

Existing approaches, primarily mask-based inpainting~\cite{lugmayr2022repaint}, operate entirely in the 2D pixel plane and have three systematic weaknesses. First, without 3D spatial awareness, modifying object coordinates in the image domain violates perspective constraints and produces distorted geometry. Second, the strong generative prior of diffusion models tends to hallucinate remnants of the original object at its old position, a phenomenon called ghosting. Third, forcing structural proxies such as hard masks into the latent space introduces a distribution mismatch: abrupt interventions break the variance homogeneity that Diffusion Transformers~\cite{peebles2023dit} assume, and the resulting self-attention leakage causes background blur and structural collapse.

These failures share a common root: the edited foreground and the uncovered background impose asymmetric generative demands. The relocated object must rigidly follow a prescribed 3D geometry, while the exposed background, which now contains disoccluded holes and the original object's footprint, needs generative freedom to synthesize plausible content. No single denoising trajectory can satisfy both at once.

We introduce GeoEdit, a framework that resolves this conflict through a principled \emph{Lift-Manipulate-Render-Denoise} pipeline. Inspired by decoupled 3D reconstruction approaches~\cite{cao2026freeorbit4d}, our lifting phase independently recovers the global scene and a geometry-complete foreground object, then registers them through cross-space point matching into a unified coordinate frame. Users apply precise spatial manipulations, and the manipulated object is rendered back into a geometry-aligned proxy image alongside a structural depth map that provides rigid conditioning for generation.

To turn this coarse proxy into a photorealistic composite, we propose Dual-Branch Denoising. We repurpose a video diffusion backbone~\cite{wan2025wan} whose temporal prior inherently preserves object identity; no identity-specific fine-tuning is needed. Motivated by the region-dependent scheduling of Time-to-Move~\cite{singer2025time}, we inject the 3D proxy into the foreground region only within a carefully selected denoising window. The injection uses variance-homogeneous noise matching so that the modified latent is statistically indistinguishable from the native denoising path; this prevents self-attention leakage. Outside the window, the generative prior freely removes the original object and fills in the background. The result is an asymmetric treatment of foreground rigidity and background flexibility, without architectural changes or additional training.

The main contributions of this work are:
\begin{itemize}
    \item A geometry-aware training-free editing framework based on a \emph{Lift-Manipulate-Render-Denoise} pipeline that lifts editing into 3D for physically plausible object manipulation, without external 3D software~\cite{chen2025blenderfusion} or synthetic training data~\cite{michel2023object}.
    \item Dual-Branch Denoising, which uses a video diffusion backbone for identity preservation and injects 3D constraints via variance-homogeneous injection within a selective denoising window, resolving the asymmetric constraint between foreground rigidity and background flexibility.
    \item GeoEditBench, a benchmark covering object translation, object rotation, and camera movement with pose-aware evaluation metrics, along with consistent improvements over existing methods.
\end{itemize}
\begin{figure}[h]
    \centering
    \includegraphics[width=\linewidth]{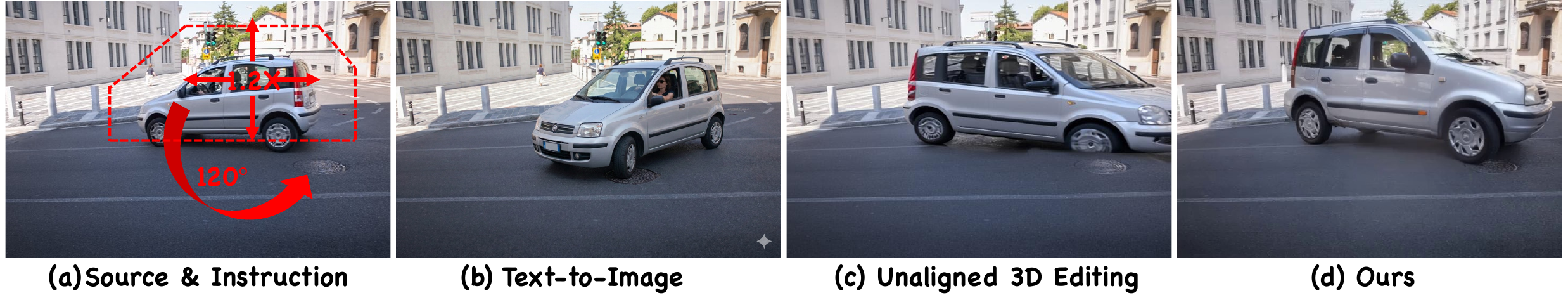}
    \caption{\textbf{Comparison with previous approaches on geometry-aware object manipulation.} 
Given a source image and an instruction requiring a $120^\circ$ rotation and $1.2\times$ scaling, existing methods struggle to maintain geometric consistency. 
In contrast, our method faithfully follows the specified transformation while preserving object identity and producing coherent, realistic results.}
    \label{fig:gallery}
\end{figure}

\section{Related Work}
\label{sec:02_relatedworks}

\noindent \textbf{Image editing with diffusion models.}
GAN-based models~\cite{karras2019style,isola2017image} first enabled controllable editing through latent manipulation~\cite{shen2020interpreting} and text-driven synthesis~\cite{patashnik2021styleclip}. Diffusion models~\cite{ho2020denoising,rombach2022high,ma2026group,ma2024followpose,ma2025followcreation,ma2026fastvmt,ma2025followyourmotion,ma2025controllable,ma2025followfaster,ma2024followyouremoji,ma2025followyourclick,shen2025follow,chen2025contextflow,xu2026smrabooth} have since become the dominant paradigm: InstructPix2Pix~\cite{brooks2023instructpix2pix} supports text-guided editing, and RePaint~\cite{lugmayr2022repaint} enables mask-based inpainting. ControlNet~\cite{zhang2023controlnet} adds spatial conditions such as depth~\cite{ranftl2021vision} and edges~\cite{xie2015holisticallynestededgedetection}, while DragGAN~\cite{pan2023drag} and DragDiffusion~\cite{shi2024dragdiffusion} allow point-based manipulation. These methods all operate in the 2D image plane and struggle with out-of-plane transformations that require 3D spatial reasoning. Similarly, recent 2D editing and insertion methods such as FateZero~\cite{fatezero}, Ctrl\&Shift~\cite{ctrl_shift}, and ObjectAdd~\cite{zhang2025objectadd} provide useful semantic or composition control but lack explicit physical constraints for prescribed 3D transformations.

\noindent \textbf{Geometry-aware object editing.}
Single-image lifting methods~\cite{liu2023zero1to3zeroshotimage3d,liu2023syncdreamer,he2024id,he2025fulldit2,ye2025unic,zhao2026dydit,wang2024taming, feng2025dit4edit, wang2024cove, ma2025magicstick, yang2025unified, dynamictuning_rapid,nan2026accelerating} synthesize novel views but often lack appearance consistency when composited into real scenes. Dataset-driven approaches~\cite{chang2015shapenet,deitke2023objaverse} suffer from synthetic-to-real domain gaps~\cite{yang2023contranerf,wiles2020synsin}. Among diffusion-based works, GeoDiffuser~\cite{sajnani2025geodiffuser} and Diffusion Handles~\cite{pandey2024diffusion} inject 3D conditions, while Object-3DIT~\cite{michel2023object} uses language guidance but relies on synthetic data. ObjectMover~\cite{yu2025objectmover} uses video priors for translation, and BlenderFusion~\cite{chen2025blenderfusion} requires external 3D software. Recently, 3DitScene~\cite{3ditscene} lifts pixels to 3D Gaussians for Score Distillation Sampling (SDS) editing. While semantic alignment is competitive, SDS incurs high computational costs and yields weaker geometric correctness. In contrast, our training-free pipeline uses monocular lifting to support full rigid transformations (rotation, scaling) without external software or synthetic data.

\noindent \textbf{Denoising strategies for structured editing.}
Injecting structural constraints into pre-trained diffusion models without compromising generation quality remains a core challenge. SDEdit~\cite{meng2021sdedit} balances fidelity and realism via timestep-based initialization, while Blended Diffusion~\cite{avrahami2022blended} and DiffEdit~\cite{couairon2022diffedit} rely on masked blending, often causing boundary artifacts or distribution mismatch. More recently, Time-to-Move~\cite{singer2025time} explores region-dependent scheduling for video generation. Our Dual-Branch Denoising resolves the asymmetric tension between rigid foreground geometry and generative background freedom via variance-homogeneous injection within a selective denoising window, which preserves latent statistics and effectively suppresses self-attention leakage~\cite{huang2026exposurebiasalleviatedirectional}.

\section{Method}

\begin{figure*}[t]
    \centering
    \includegraphics[width=\linewidth]{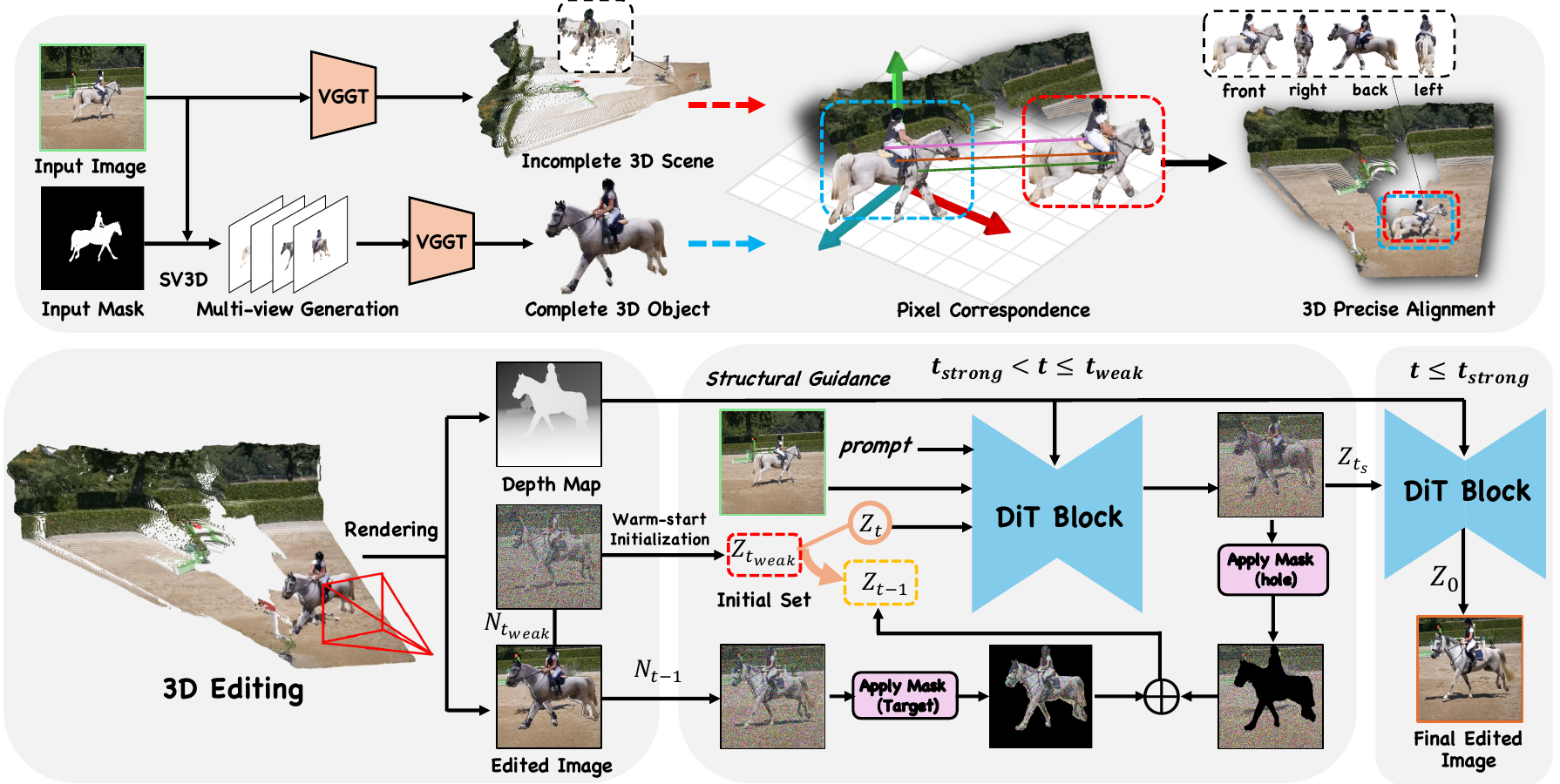}
    \caption{\textbf{Overview of the proposed framework.} Top: Decoupled 3D reconstruction and precise alignment pipeline. Bottom: Dual-branch denoising architecture featuring warm-start initialization and variance-homogeneous injection.}
    \label{fig:framework}
\end{figure*}

We present a geometry-aware framework for precise object manipulation. As shown in Fig.~\ref{fig:framework}, our pipeline follows a 2D$\to$3D$\to$2D workflow. While the 3D lifting and rendering modules leverage existing techniques, the core contribution of GeoEdit is a dual-branch denoising architecture with variance-homogeneous injection. This mechanism mitigates attention leakage by aligning injected noise statistics with the pre-trained diffusion backbone, enabling rigid foreground transformation alongside free-form background synthesis. The source image is first lifted into a 3D representation for user-defined spatial transformations, rendered into a 2D proxy ($I_\text{proxy}$) with a depth map, and finally converted into a photorealistic composite via our dual-branch denoiser (Sec.~\ref{sec:lifting},~\ref{sec:denoising}).

\vspace{1mm}
\noindent\textbf{Problem Formulation.} Given a source image $I_\text{src} \in \mathbb{R}^{H \times W \times 3}$ and a user-specified 3D transformation $\mathcal{T} = \{\mathbf{R}, \mathbf{t}, s\}$ on a target object, the framework produces geometrically aligned conditioning signals: a geometry-aligned proxy $I_\text{proxy} \in \mathbb{R}^{H \times W \times 3}$, a structural depth map $D_\text{rep} \in \mathbb{R}^{H \times W}$, and a spatial mask $M \in \{0, 1\}^{H \times W}$ obtained by projecting the transformed object silhouette back into the image plane. The goal is to generate a realistic composite $x_0 \in \mathbb{R}^{H \times W \times 3}$ that strictly respects $\mathcal{T}$ while preserving the unedited background of $I_\text{src}$.

\subsection{Preliminaries: Latent Diffusion Models}
\label{sec:preliminaries}

Latent Diffusion Models (LDMs)~\cite{ho2020denoising, rombach2022high} operate in a compressed latent space to reduce computational cost while maintaining generation quality. Given an RGB image $x \in \mathbb{R}^{H \times W \times 3}$, a pre-trained encoder maps it to a latent representation $z_0 = \mathcal{E}(x)$.
The forward diffusion process gradually adds Gaussian noise to $z_0$ over $T$ steps. At timestep $t \in [1, T]$, the noisy latent $z_t$ is:
\begin{equation}
    z_t = \sqrt{\bar{\alpha}_t}\, z_0 + \sqrt{1 - \bar{\alpha}_t}\, \epsilon, \quad \epsilon \sim \mathcal{N}(0, \mathbf{I})
\end{equation}
where $\bar{\alpha}_t$ follows a predefined noise schedule. During the reverse process, a network $\epsilon_\theta$, parameterized as a UNet or Diffusion Transformer (DiT), is trained to predict the added noise $\epsilon$ given $t$ and optional conditioning signals $c$:
\begin{equation}
    L_{LDM} = \mathbb{E}_{z_0, \epsilon \sim \mathcal{N}(0, \mathbf{I}), t, c} \left[ \|\epsilon - \epsilon_\theta(z_t, t, c)\|_2^2 \right]
\end{equation}
Once sampling is complete, a decoder maps the denoised latent back to pixel space: $x' = \mathcal{D}(z_0)$. Our method does not fine-tune any model weights; it intervenes in the reverse sampling process to inject 3D geometric constraints into a pre-trained DiT.

\subsection{Decoupled 3D Reconstruction and Canonical Alignment}
\label{sec:lifting}

To overcome the occlusion ambiguity inherent in monocular lifting, we reconstruct the global scene and the target object independently.

\vspace{1mm}
\noindent\textbf{Decoupled 3D Reconstruction.} As shown in Fig.~\ref{fig:framework}, we first use a monocular depth estimator~\cite{wang2025vggt} to lift the source image $I_\text{src}$ into an incomplete 3D scene. The spatial mask $M$ segments this scene into a background $\mathcal{P}_{bg}$ and a visible foreground anchor $\mathcal{P}_{fg}^\text{vis}$. To enable precise manipulation, the occluded surfaces of the target object must be recovered. We apply a multi-view diffusion model~\cite{voleti2024sv3d} to the masked foreground to synthesize novel views, constructing a geometry-complete object point cloud $\mathcal{P}_{fg}^\text{comp}$ in an isolated canonical space.

\vspace{1mm}
\noindent\textbf{Correspondence-Aware Alignment.} To perform physically plausible edits, the isolated object must be registered back into the global scene. Both $\mathcal{P}_{fg}^\text{comp}$ and $\mathcal{P}_{fg}^\text{vis}$ originate from $I_{src}$, so points at the same pixel coordinate $\mathbf{u}$ correspond to the same physical surface point~\cite{cao2026freeorbit4d}. We use this to extract dense 3D--3D correspondences:
\begin{equation}
    \mathcal{C} = \left\{ \left( \mathcal{P}_{fg}^\text{comp}(\mathbf{u}),\, \mathcal{P}_{fg}^\text{vis}(\mathbf{u}) \right) \mid M(\mathbf{u}) = 1 \right\}
\end{equation}
From the correspondence set $\mathcal{C}$, we compute the optimal similarity transformation (rotation $\mathbf{R}_\text{align}$, translation $\mathbf{t}_\text{align}$, scale $s_\text{align}$) by minimizing the registration error:
\begin{equation}
    \mathop{\arg\min}\limits_{\mathbf{R}_\text{align},\, \mathbf{t}_\text{align},\, s_\text{align}} \sum_{(\mathbf{p}, \mathbf{q}) \in \mathcal{C}} \left\| s_\text{align}\, \mathbf{R}_\text{align}\, \mathbf{p} + \mathbf{t}_\text{align} - \mathbf{q} \right\|_2^2
\end{equation}
where $\mathbf{p} = \mathcal{P}_{fg}^\text{comp}(\mathbf{u})$ and $\mathbf{q} = \mathcal{P}_{fg}^\text{vis}(\mathbf{u})$ denote the matched 3D points. This aligns $\mathcal{P}_{fg}^\text{comp}$ into the global scene space, establishing a unified coordinate frame in which the user can precisely apply the desired manipulation $\mathcal{T}$.

\vspace{1mm}
\noindent\textbf{Proxy Rendering.} To render the final conditioning signals, the original visible foreground is removed, leaving an exposed background hole in $I_{src}$. Instead of relying on a black-box heuristic, this hole is efficiently filled using the Telea inpainting algorithm~\cite{telea2004image} based on the fast marching method, providing a coarse but structurally continuous global color layout. The aligned and manipulated 3D object is then re-projected over this completed background to form the geometry-aligned proxy $I_{proxy}$ and the structural depth map $D_{rep}$. While the background texture is coarse at this stage, the downstream denoising uses it as a structural baseline and synthesizes realistic high-frequency details.

\subsection{Dual-Branch Denoising and Variance Injection}
\label{sec:denoising}

\begin{figure}[t]
    \centering
    \includegraphics[width=\linewidth]{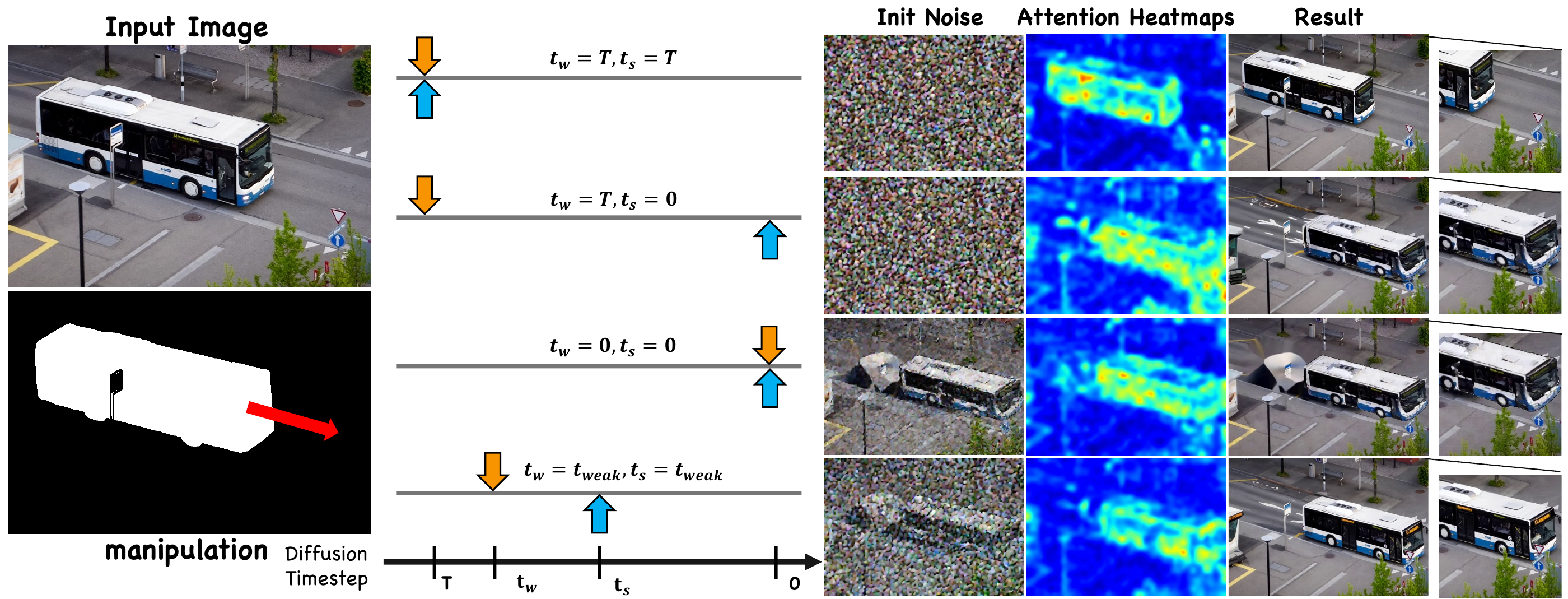}
    \caption{\textbf{Visualizing the generative trade-off.} Different configurations of initialization ($t_\text{weak}$) and injection ($t_\text{strong}$) timesteps dictate whether the model leans towards preserving the rigid object skeleton or hallucinating semantic background details.}
    \label{fig:attention}
\end{figure}

After the 3D editing and rendering process, the geometry-aligned proxy $I_\text{proxy}$ has correct 3D structure but coarse textures and exposed background holes. Adapting this proxy via standard single-timestep initialization (SDEdit~\cite{meng2021sdedit}) presents a trade-off between structural preservation and semantic realism. As the attention heatmaps in Fig.~\ref{fig:attention} illustrate, a large initialization timestep affords the generative prior enough freedom to hallucinate realistic backgrounds, but inevitably degrades the prescribed 3D object skeleton, causing severe geometric drift. Conversely, a small timestep forces strict adherence to the proxy but retains coarse, unrealistic background artifacts.

We argue that the manipulated foreground and the unedited background require asymmetric generative constraints: the target object demands rigid adherence to the 3D spatial signal, whereas the background requires generative flexibility. Furthermore, standard DiTs assume a globally homogeneous noise distribution. Directly forcing mismatched latents into specific regions disrupts this homogeneity, causing self-attention leakage and global blurring. To reconcile these constraints, we introduce a training-free \textit{Dual-Branch Denoising} strategy (Fig.~\ref{fig:framework}, bottom).

\vspace{1mm}
\noindent\textbf{Video Diffusion Backbone.} We build on a depth-conditioned video diffusion model $\Phi_{\text{VDM}}$~\cite{wan2025wan, jiang2025vace} as our generation backbone. The source image $I_\text{src}$ is the appearance reference, $D_\text{rep}$ provides structural depth conditioning through ControlNet-style control blocks~\cite{zhang2023controlnet} that inject geometric features into the transformer, and a text prompt $c$ guides semantic context:
\begin{equation}
    x_{0} = \Phi_{\text{VDM}} \left( I_\text{src}, D_\text{rep}, c \right) \big|_{\text{last frame}}
\end{equation}
Depth maps act as a geometric scaffold that anchors the object's skeleton against degradation during high-noise phases. To repurpose this video model for single-image editing, we construct a short target sequence by repeating the proxy as identical frames; the temporal self-attention of $\Phi_{\text{VDM}}$ then enforces cross-frame consistency, which naturally preserves the object's identity without any identity-specific fine-tuning or adapters. The last generated frame is taken as the output.


\begin{algorithm}[t]
\caption{Dual-Branch Denoising}
\label{alg:dual_branch}
\begin{algorithmic}[1]
\REQUIRE Source image $I_{\text{src}}$, proxy $I_{\text{proxy}}$, depth $D_{\text{rep}}$, mask $M$, window bounds $t_{\text{weak}}$, $t_{\text{strong}}$, video backbone $\Phi_{\text{VDM}}$
\ENSURE  Edited composite $x_0$
\STATE $z_{\text{proxy}} \gets \mathcal{E}(I_{\text{proxy}})$ \COMMENT{Encode proxy to latent space}
\STATE $\epsilon' \sim \mathcal{N}(0, \mathbf{I})$ \COMMENT{Sample fixed noise (reused across all steps)}
\STATE $z_{t_{\text{weak}}} \gets \sqrt{\bar{\alpha}_{t_{\text{weak}}}}\, z_{\text{proxy}} + \sqrt{1 - \bar{\alpha}_{t_{\text{weak}}}}\, \epsilon'$ \COMMENT{Warm-start initialization}
\FOR{$t = t_{\text{weak}}, \dots, 1$}
    \STATE $z_{t-1}^{\text{pred}} \gets \text{Denoise}(\Phi_{\text{VDM}},\, z_t,\, t,\, I_{\text{src}},\, D_{\text{rep}})$ \COMMENT{Backbone step}
    \IF{$t_{\text{strong}} < t \le t_{\text{weak}}$}
        \STATE $z_{t-1}^{\text{ref}} \gets \sqrt{\bar{\alpha}_{t-1}}\, z_{\text{proxy}} + \sqrt{1 - \bar{\alpha}_{t-1}}\, \epsilon'$ \COMMENT{Variance-matched proxy}
        \STATE $z_{t-1} \gets M \odot z_{t-1}^{\text{ref}} + (1-M) \odot z_{t-1}^{\text{pred}}$ \hfill \COMMENT{Replace foreground only}
    \ELSE
        \STATE $z_{t-1} \gets z_{t-1}^{\text{pred}}$ \COMMENT{Outside window: free running}
    \ENDIF
\ENDFOR
\STATE $x_0 \gets \mathcal{D}(z_0)$ \COMMENT{Decode to pixel space}
\end{algorithmic}
\end{algorithm}
\vspace{1mm}
\noindent\textbf{Warm-Start Initialization.} We avoid the semantic ambiguity of pure noise ($t=T$) by initializing the reverse process at an intermediate timestep $t_\text{weak}$. We encode $I_\text{proxy}$ and perturb it to $t_\text{weak}$:
\begin{equation}
    z_{t_\text{weak}} = \sqrt{\bar{\alpha}_{t_\text{weak}}}\, \mathcal{E}(I_\text{proxy}) + \sqrt{1 - \bar{\alpha}_{t_\text{weak}}}\, \epsilon', \quad \epsilon' \sim \mathcal{N}(0, \mathbf{I})
\end{equation}
This establishes a global color layout and structural baseline, so the model retains the scene's overall identity from the start of the denoising trajectory.

\vspace{1mm}
\noindent\textbf{Variance-Homogeneous Injection.} To enforce asymmetric constraints, we perform selective latent replacement within a denoising window $t_\text{strong} < t \le t_\text{weak}$. At each step $t$ within this window, instead of raw feature replacement, we synchronize the masked object region with the forward-noised proxy~\cite{singer2025time}. Building on the masked blending of Blended Diffusion~\cite{avrahami2022blended} but strictly maintaining variance homogeneity, we formulate the injection as:
\begin{equation}
\begin{split}
    \tilde{z}_{t-1} = & M \odot \left( \sqrt{\bar{\alpha}_{t-1}}\, \mathcal{E}(I_\text{proxy}) + \sqrt{1 - \bar{\alpha}_{t-1}}\, \epsilon' \right) \\
    & + (1-M) \odot z_{t-1}^\text{pred}
\end{split}
\end{equation}
where $z_{t-1}^\text{pred}$ is the latent predicted by the $\Phi_{\text{VDM}}$ backbone, and $\epsilon' \sim \mathcal{N}(0, \mathbf{I})$ is a fixed Gaussian noise tensor sampled once at the start. The foreground thus tracks the proxy's 3D geometry, while the background $z_{t-1}^\text{pred}$ evolves under the generative prior. Because the injected signal has exactly the variance that the noise schedule prescribes at timestep $t{-}1$, it is statistically indistinguishable from the native denoising path, and the self-attention mechanism sees a spatially homogeneous distribution. To quantitatively validate this, we introduce the Attention Leakage Ratio (ALR) to measure the fraction of background-query attention mass assigned to foreground proxy tokens, defined as $\mathrm{ALR}{=}\mathrm{mean}_{l,h} A^{l,h}_{\mathcal{B}\rightarrow\mathcal{F}}/A^{l,h}_{\mathcal{B}\rightarrow *}$. Our analysis confirms that this variance-matched injection effectively suppresses self-attention leakage, notably reducing the ALR from 8.4\% to 6.4\% at the peak leakage step ($t=46$). Fixing $\epsilon'$ across all steps maintains structural consistency of the injected proxy. The complete dual-branch denoising procedure is summarized in Algorithm~\ref{alg:dual_branch}.

\vspace{1mm}
\noindent\textbf{Global Harmonization.} Once the denoising crosses $t \le t_\text{strong}$, the mask injection loop terminates. The latent evolves freely under the pre-trained generative prior. During this phase, the model harmonizes mask boundaries and synthesizes high-frequency textures to fill disoccluded holes. The final decoded frame is the output composite.

\section{Experiments}

\subsection{Implementation Details}
\label{sec:implementation}
Our framework operates as a fully training-free pipeline, strategically orchestrating off-the-shelf pre-trained models across all intermediate stages. Specifically, we adopt VGGT \cite{wang2025vggt} for dense 3D point-map reconstruction from both the source image and the synthesized multi-view observations. To achieve canonical object completion, SV3D \cite{voleti2024sv3d} is utilized to generate geometrically consistent novel views of the isolated foreground. The terminal dual-branch denoising phase is powered by Wan2.2-VACE \cite{wan2025wan, jiang2025vace}, which serves as our depth-conditioned generative prior. 

All experiments are conducted on a single NVIDIA A800 GPU at a standardized resolution of $720 \times 480$ pixels. For variance-homogeneous injection, we use a diffusion schedule of $T=50$ timesteps, setting the warm-start threshold to $t_\text{weak}=47$ and ending mask injection at $t_\text{strong}=40$. This empirically determined window achieves the best trade-off between geometric fidelity and background harmonization. Runtime analysis shows that the 3D extraction stage requires approximately 2 minutes and 19 GB of VRAM, whereas the denoising stage takes about 16 minutes and 44 GB of VRAM.

\begin{figure*}[t]
  \centering
  \includegraphics[width=\linewidth]{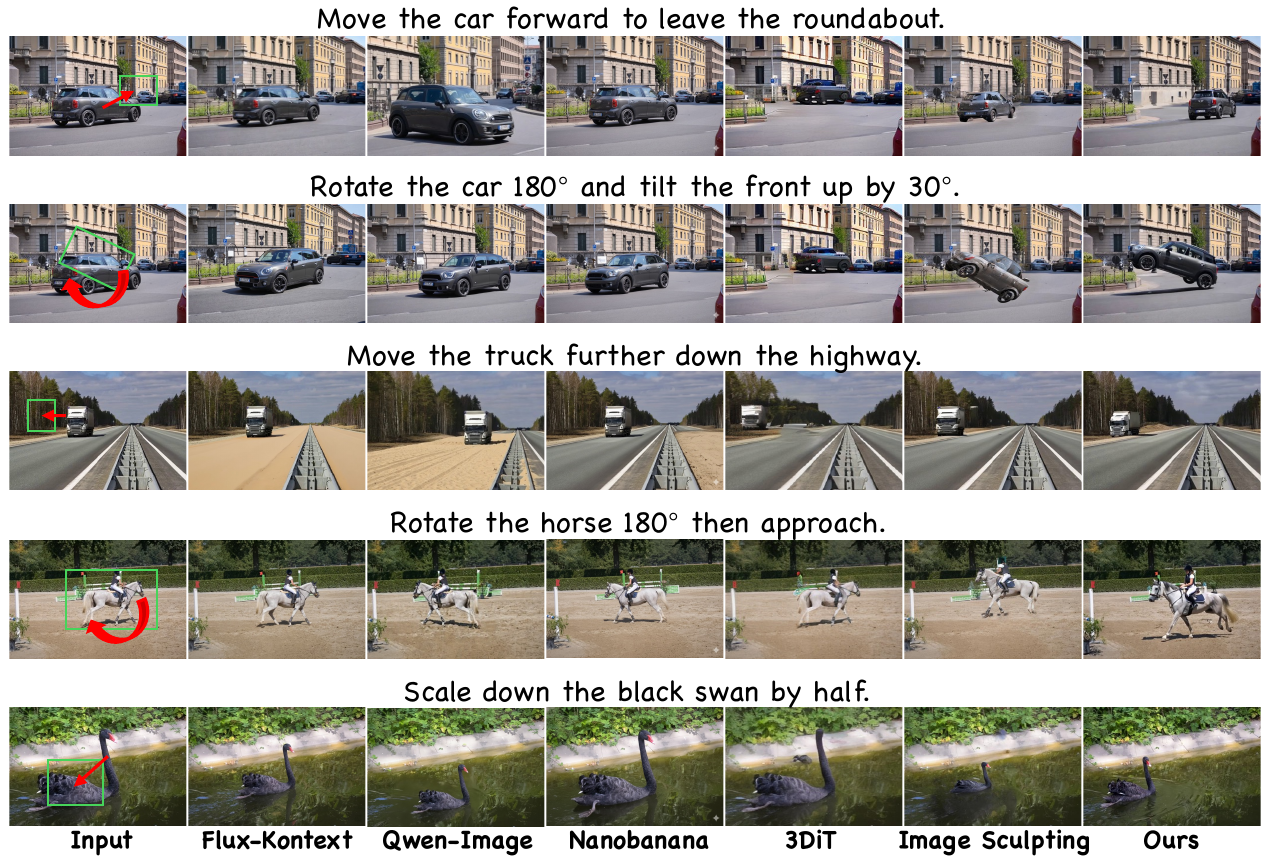}
  \caption{\textbf{Qualitative comparison of different methods on object manipulation tasks}. Our model achieves superior performance compared to state-of-the-art methods in background preservation and geometric consistency. }
  \label{fig:comparison}
\end{figure*}

\subsection{GeoEditBench}

To systematically evaluate geometry-aware image editing, we introduce GeoEditBench, a benchmark of 200 image pairs collected from diverse real-world scenarios. We curate the dataset by removing samples with ambiguous geometry, severe rendering artifacts, or inconsistent illumination. The benchmark is divided into three categories according to the primary spatial transformation: 80 \textbf{object translation} examples, 80 \textbf{object rotation} examples, and 40 \textbf{camera movement} examples. This category-aware design enables both overall and per-category evaluation of spatial editing performance under diverse geometric manipulations.

\subsection{Comparison with Baselines}

\noindent\textbf{Quantitative comparison.}
Table~\ref{tab:comparison_2} reports the quantitative zero-shot evaluation of our method against several recent baselines on GeoEditBench, evaluated on the same pre-defined fixed 50-pair stratified subset used throughout Tabs.~\ref{tab:comparison_2}-\ref{tab:ablation_components}. Following prior work, we evaluate performance across three dimensions: (i) reconstruction fidelity using PSNR; (ii) object identity preservation using DINO~\cite{caron2021dino} and CLIP similarity~\cite{radford2021clip}; and (iii) perceptual discrepancy using LPIPS and DreamSim~\cite{fu2023dreamsim}. To assess the physical correctness of 3D manipulations, we further introduce PoseMap IoU and Object IoU. PoseMap IoU compares predicted and target pose maps, while Object IoU measures silhouette alignment with the transformed target mask; both rely on external estimators independent of GeoEdit. Overall, our method achieves the strongest performance on most metrics, obtaining the highest PSNR (23.499) and DINO (0.961) alongside the lowest LPIPS (0.114) and DreamSim (0.027). On the geometry-aware metrics, GeoEdit further achieves 94.9\% PoseMap IoU and 57.9\% Object IoU, demonstrating precise geometric control. Although NanoBanana attains a slightly higher CLIP score (0.976 vs. 0.952), our method provides the strongest overall balance between geometric correctness, identity preservation, perceptual quality, and background fidelity. Additional evaluations (3DEdit-Bench, per-category results, and confidence intervals) are provided in the Supplementary Material.


\begin{table*}[t] 
\centering
\setlength{\tabcolsep}{4.5pt} 

\caption{Comparison with state-of-the-art object editing methods on the fixed 50-pair stratified subset of GeoEditBench. Preference scores (Pref. $\uparrow$) indicate VLM (AI) and human ratings. Geometry-aware metrics explicitly evaluate 3D transformation correctness. \textcolor{red}{Red} and \textcolor{blue}{Blue} denote the best and second best results.}
\label{tab:comparison_2}

\resizebox{0.995\linewidth}{!}{
\begin{tabular}{l|ccccccc|cc}
\toprule
\multirow{2}{*}{Method}
& \multicolumn{7}{c|}{Quantitative metrics}
& \multicolumn{2}{c}{Pref. $\uparrow$} \\
\cmidrule(lr){2-8}\cmidrule(lr){9-10}
& CLIP $\uparrow$ & PSNR $\uparrow$ & DINO $\uparrow$ & LPIPS $\downarrow$ & DreamSim $\downarrow$ & PoseMap IoU $\uparrow$ & Object IoU $\uparrow$ 
& AI & Human \\
\midrule
Qwen-Image~\cite{wu2025qwenimagetechnicalreport}
& 0.885 & 14.013 & 0.685 & 0.431 & 0.185 & 66.7\% & 8.8\% & 2.010 & 2.667 \\

NanoBanana~\cite{nanobanana}
& \textbf{\textcolor{red}{0.976}} & 19.164 & \textbf{\textcolor{blue}{0.948}} & \textbf{\textcolor{blue}{0.118}} & \textbf{\textcolor{blue}{0.031}} & 75.0\% & 25.2\% & 3.085 & 2.286 \\

\midrule

3DiT~\cite{michel2023object}
& 0.941 & 21.532 & 0.771 & 0.371 & 0.084 & 60.0\% & 33.5\% & 1.300 & 1.191 \\

Flux-Kontext~\cite{labs2025flux}
& 0.927 & 20.250 & 0.841 & 0.261 & 0.112 & 66.7\% & \textbf{\textcolor{blue}{51.3\%}} & 2.563 & 2.714 \\

Image Sculpting~\cite{yenphraphai2024image}
& 0.939 & \textbf{\textcolor{blue}{22.101}} & 0.802 & 0.147 & 0.104 & \textbf{\textcolor{blue}{80.0\%}} & 28.2\% & 2.215 & 2.619 \\
\midrule
\textbf{Ours}
& \textbf{\textcolor{blue}{0.952}} & \textbf{\textcolor{red}{23.499}} & \textbf{\textcolor{red}{0.961}} & \textbf{\textcolor{red}{0.114}} & \textbf{\textcolor{red}{0.027}} & \textbf{\textcolor{red}{94.9\%}} & \textbf{\textcolor{red}{57.9\%}} & \textbf{\textcolor{red}{4.125}} & \textbf{\textcolor{red}{4.810}} \\
\bottomrule
\end{tabular}
}
\end{table*}

\begin{table}[t]
  \centering
  \caption{\textbf{Quantitative Ablation on Timestep Thresholds}. We evaluate the trade-off between geometric structure preservation and semantic appearance generation. Our configured interval $(t_w=47, t_s=40)$ achieves the optimal balance across all metrics.}
  \label{tab:ablation_timesteps}
  \renewcommand{\arraystretch}{1.3} 
  \setlength{\tabcolsep}{3pt} 
  \resizebox{\linewidth}{!}{
  \begin{tabular}{lcccc}
  \toprule
  Configuration $(t_w, t_s)$ & PSNR $\uparrow$ & DINO $\uparrow$ & CLIP $\uparrow$ & DreamSim $\downarrow$ \\
  \midrule
  Pure Prior (50, 50)       & 18.479 & 0.894 & 0.921 & 0.057 \\
  SDEdit Init (47, 47)      & 20.724 & 0.932 & 0.921 & 0.044 \\
  Low Noise Init (40, 40)   & 21.822 & 0.942 & 0.906 & 0.060 \\
  Full Injection (50, 0)    & 20.012 & 0.860 & 0.871 & 0.050 \\
  Pure Proxy (1, 1)         & 19.465 & 0.812 & 0.806 & 0.084 \\
  \midrule
  \textbf{Ours Optimal (47, 40)} & \textbf{23.499} & \textbf{0.961} & \textbf{0.952} & \textbf{0.027} \\
  \bottomrule
  \end{tabular}%
  } 
\end{table}

\noindent\textbf{Qualitative comparison.}
Figure~\ref{fig:comparison} presents a qualitative comparison between our method and several state-of-the-art approaches, including Image Sculpting~\cite{yenphraphai2024image}, NanoBanana~\cite{nanobanana}, Flux-Kontext~\cite{labs2025flux}, Qwen-Image-Edit~\cite{wu2025qwenimagetechnicalreport}, and 3DiT~\cite{michel2023object}. 3DiT exhibits limited generalization to real-world data, likely due to its reliance on specific training datasets. Image Sculpting demonstrates relatively strong object manipulation capability; however, it tends to lose fine-grained details, which affects overall visual fidelity. Flux-Kontext and Qwen-Image-Edit show some limitations in maintaining background consistency, leading to noticeable inconsistencies in certain surrounding regions after editing. NanoBanana preserves background content comparatively well, but its ability to perform precise object manipulation appears less robust in complex spatial adjustments. In contrast, our method achieves more reliable camera pose control and object relocation while maintaining consistent background appearance, resulting in visually coherent and geometrically aligned outputs.


\vspace{1mm}
\noindent\textbf{Human and AI Preference Evaluation.} 
Given the inherently subjective nature of generative editing, we conduct comprehensive qualitative assessments utilizing both human participants and an advanced Vision-Language Model (VLM). We evaluate the generated results based on three core criteria: (1) \textit{Object Identity Preservation}, assessing whether the manipulated target strictly retains its original appearance; (2) \textit{Instruction Fidelity}, measuring how accurately the visual edit reflects the provided spatial command; and (3) \textit{Background Preservation}, evaluating the consistency and intactness of the unedited regions. 

For the human preference study, we collected questionnaire responses from 25 independent participants. Evaluators were presented with the original source image alongside the specific editing instruction, and were asked to rate the generated outputs from different models on a scale ranging from 0 to 5, where higher scores indicate superior generation quality. The final human preference score for each method is computed as the average rating across all participants and test cases, achieving a Krippendorff's $\alpha$ of 0.74 which indicates substantial inter-rater agreement.

To complement the human study and provide a scalable complementary assessment, we employ a Gemini-family multimodal model~\cite{team2023gemini} as an expert VLM judge. Prompted to act as an impartial image quality scorer, the VLM is provided with the identical input trio—the source image, the manipulation instruction, and the generated result. It analyzes visual coherence and instruction alignment to assign a comprehensive quality score from 0 to 5. As reported in Tab. \ref{tab:comparison_2}, our framework secures the highest preference scores from both human evaluators and the AI judge, corroborating its robust perceptual superiority (inter-rater agreement details in Suppl. Material).

\begin{figure}[t] 
    \centering
    \includegraphics[width=\linewidth]{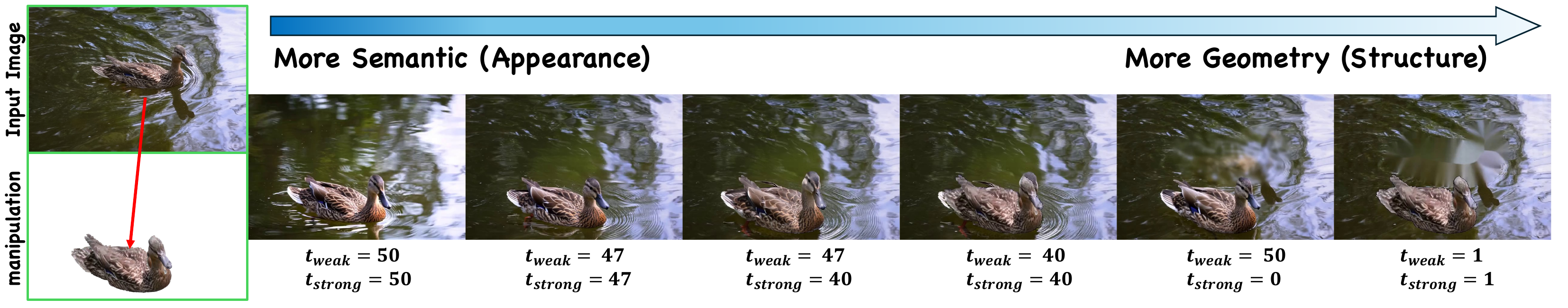}
    \caption{\textbf{Visual Ablation on Timestep Thresholds}. We illustrate the fundamental trade-off between semantic realism (left) and geometric structure preservation (right). Relying predominantly on the generative prior (e.g., $t_w=50, t_s=50$) grants excessive freedom, resulting in structural deviation from the proxy. Conversely, excessive proxy injection at low noise levels (e.g., $t_w=1, t_s=1$) rigidly preserves geometry but introduces severe rendering artifacts and blending failures. Our optimal configuration ($t_w=47, t_s=40$) strikes the perfect balance, achieving seamless background harmonization while strictly adhering to the intended 3D spatial transformation.}
    \label{fig:ablation_timesteps_visual}
\end{figure}

\begin{figure}[t]
    \centering
    \includegraphics[width=\linewidth]{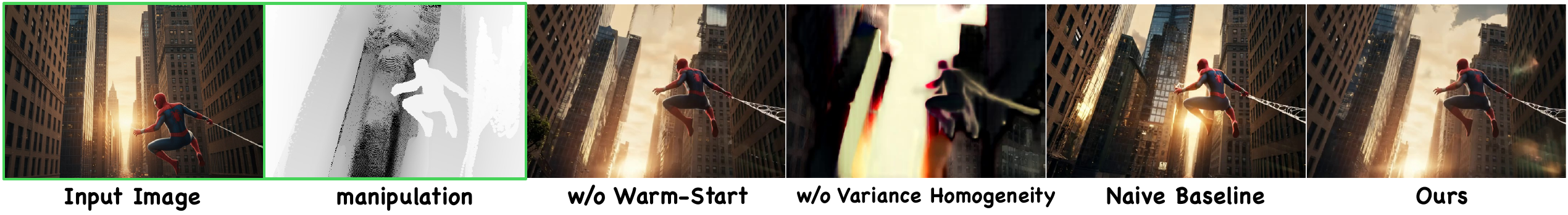}
    \caption{\textbf{Qualitative Ablation on Proposed modules}. We demonstrate the visual impact of each core module. The Naive Baseline struggles with both 3D skeleton preservation and background fidelity, yielding a distorted pose and altered context. Removing the warm-start initialization (w/o Warm-Start) results in an unnatural background synthesis, failing to smoothly harmonize the generated textures with the original scene. Critically, injecting uncalibrated spatial constraints (w/o Variance Homogeneity) severely disrupts the homogeneous latent distribution, leading to catastrophic rendering artifacts. In contrast, Ours seamlessly executes the complex spatial manipulation while preserving a highly natural and faithful background.}
    \label{fig:ablation_visual_components}
\end{figure}

\subsection{Ablation Studies} 
\vspace{1mm}

\noindent\textbf{Effectiveness of Dual-Branch Denoising.}
To analyze the effectiveness of our dual-branch denoising framework, we conduct ablation studies on the critical timestep thresholds summarized in Tab.~\ref{tab:ablation_timesteps}. The results reveal a clear trade-off between geometric structure preservation and semantic realism under conventional editing settings. As shown in Fig.~\ref{fig:ablation_timesteps_visual}, relying on single-timestep initialization struggles to balance foreground geometry and background generation. A high initialization noise at step 47 increases generative freedom and achieves a CLIP score of 0.921, but weakens geometric fidelity, leading to a lower PSNR of 20.724 and a DreamSim error of 0.044. Reducing the initialization noise to step 40 improves spatial alignment (PSNR 21.822) but limits semantic diversity, lowering the CLIP score to 0.906.

Extreme configurations further highlight this limitation. Initializing from pure noise destroys the prescribed geometry (PSNR 18.479), while excessive proxy injection exposes rendering artifacts and degrades feature consistency (DINO 0.812, DreamSim 0.084). Moreover, enforcing full injection across all timesteps disrupts the diffusion process, reducing the DINO score to 0.860.

In contrast, our calibrated denoising window ($t_w=47$, $t_s=40$) achieves the best balance between structural fidelity and semantic realism, improving PSNR to 23.499 and DINO to 0.961 while reducing DreamSim to 0.027, validating the effectiveness of our asymmetric constraint strategy.

\begin{table}[t]
  \centering
  \caption{\textbf{Quantitative Ablation on Proposed Components.} We validate our framework using a leave-one-out strategy on the unified 50-pair protocol. Removing any core component significantly degrades either geometric preservation or semantic harmony, while our full model achieves the optimal balance across all metrics.}
  \label{tab:ablation_components}
  \renewcommand{\arraystretch}{1.3}
  \setlength{\tabcolsep}{2pt}
  \resizebox{\linewidth}{!}{%
  \begin{tabular}{lccccc}
  \toprule
  Model & PSNR $\uparrow$ & DINO $\uparrow$ & CLIP $\uparrow$ & LPIPS $\downarrow$ & DreamSim $\downarrow$ \\
  \midrule
  Naive Baseline           & 16.217 & 0.835 & 0.890 & 0.253 & 0.049 \\
  w/o Warm-Start           & 21.320 & 0.943 & 0.942 & 0.134 & 0.043 \\
  w/o Variance Homogeneity & 11.494 & 0.314 & 0.682 & 0.520 & 0.240 \\
  \midrule
  \textbf{Ours (Full Model)} & \textbf{23.499} & \textbf{0.961} & \textbf{0.952} & \textbf{0.114} & \textbf{0.027} \\
  \bottomrule
  \end{tabular}%
  }
\end{table}


\noindent\textbf{Effectiveness of Variance-Homogeneous Injection.}
Variance-homogeneous injection serves as the mathematical cornerstone of our spatial constraint mechanism. As reported in Tab. \ref{tab:ablation_components}, replacing our synchronized forward-noise injection with uncalibrated constraints (w/o Variance Homogeneity) leads to a catastrophic performance collapse. Disrupting the homogeneous latent distribution of the pre-trained diffusion model severely degrades deep feature representations, causing the DINO score to plummet from 0.961 to 0.314 and the CLIP score to drop to 0.682. Structurally, this lack of variance alignment induces severe geometric distortions, increasing the DreamSim error to 0.240. This quantitative degradation is vividly reflected in Fig. \ref{fig:ablation_visual_components}, where the absence of this module produces extreme attention leakage and catastrophic rendering artifacts, highlighting that variance homogeneity is critical for stable generative editing.

\vspace{1mm}
\noindent\textbf{Effectiveness of Warm-Start Initialization.}
The warm-start initialization is crucial for anchoring the global color layout and preserving the unedited background context. When this module is removed (w/o Warm-Start), the diffusion process is forced to reconstruct the scene from pure Gaussian noise without the low-frequency guidance of the source image. Tab. \ref{tab:ablation_components} demonstrates that this omission causes a decline in pixel-level and perceptual fidelity, with PSNR dropping from 23.499 to 21.320 and the LPIPS perceptual error surging from 0.114 to 0.134. Qualitatively, as demonstrated in Fig. \ref{fig:ablation_visual_components}, failing to warm-start the latents results in an unnatural background synthesis that completely alters the original illumination and environmental context. By successfully integrating this initialization, our full model reliably executes complex spatial manipulations while maintaining a highly faithful background.
\section{Conclusion}
\label{sec:conclusion}

In this paper, we presented \textbf{GeoEdit}, a principled framework for enforcing strict 3D physical constraints, including translation, rotation, and scaling, in single-image object manipulation. We show that the core challenge in diffusion-based editing arises from the asymmetric generative requirements of the scene: geometry-constrained foreground manipulation and free-form background hallucination. To address this, GeoEdit adopts a \emph{Lift-Manipulate-Render-Denoise} pipeline that lifts 2D editing into a geometry-aware 3D space. Our Dual-Branch Denoising mechanism bridges the rendered proxy and diffusion prior through variance-consistent injection, effectively preventing attention leakage and ghosting artifacts. We further introduce \textbf{GeoEditBench}, a comprehensive benchmark for evaluating spatial transformations. Extensive experiments demonstrate that GeoEdit achieves state-of-the-art performance in geometric accuracy, identity preservation, and background harmonization.

\section{Limitations}

While GeoEdit demonstrates precise 3D-aware editing, its modular pipeline introduces several limitations. First, \textbf{error propagation}: inaccuracies in depth estimation or monocular 3D lifting can distort the reconstructed geometry, leading to structural errors in the rendered proxy that propagate through subsequent stages. Second, \textbf{extreme spatial manipulations} remain challenging. Large object translations may expose disoccluded regions beyond the capacity of the background prior, causing inpainting failures, while large rotations often require hallucinating previously unseen back-facing surfaces, which can result in blurred or inconsistent textures. Third, \textbf{computational cost}: we employ the native 81-frame context of the video backbone to fully exploit its temporal prior for rigid 3D consistency; reducing this context degrades structural integrity, as shown in the Supplementary Material. Although this improves structural integrity, it increases inference time compared with purely 2D editing approaches; future work will explore lightweight fine-tuning and more efficient architectures to reduce this requirement. Finally, accurately reproducing complex view-dependent lighting, shadows, and specular effects remains difficult because these phenomena are synthesized primarily through learned generative priors rather than explicit physical modeling.

\section{Acknowledgements}

This work was supported in part by the National Natural Science Foundation of China (Grant No. 62276154), the Natural Science Foundation of Guangdong Province (Grant No. 2024TQ08X729), the Basic Research Fund of Shenzhen City (Grant Nos. JCYJ20240813112009013 and GJHZ20240218113603006), and the Major Key Project of Peng Cheng Laboratory for Experiments and Applications (Grant No. PCL2024A08).

{
    \small
    \bibliographystyle{ieeenat_fullname}
    \bibliography{main}

@String(CVPR  = {IEEE Conf. Comput. Vis. Pattern Recog.})

@String(ICLR  = {Int. Conf. Learn. Represent.})

@String(AAAI  = {AAAI})

@String(CVPR  = {CVPR})

@String(ICLR  = {ICLR})

@inproceedings{karras2019style,
  title={A style-based generator architecture for generative adversarial networks},
  author={Karras, Tero and Laine, Samuli and Aila, Timo},
  booktitle={Proceedings of the IEEE/CVF conference on computer vision and pattern recognition},
  pages={4401--4410},
  year={2019}
}

@inproceedings{isola2017image,
  title={Image-to-image translation with conditional adversarial networks},
  author={Isola, Phillip and Zhu, Jun-Yan and Zhou, Tinghui and Efros, Alexei A},
  booktitle={Proceedings of the IEEE conference on computer vision and pattern recognition},
  pages={1125--1134},
  year={2017}
}

@inproceedings{patashnik2021styleclip,
  title={Styleclip: Text-driven manipulation of stylegan imagery},
  author={Patashnik, Or and Wu, Zongze and Shechtman, Eli and Cohen-Or, Daniel and Lischinski, Dani},
  booktitle={Proceedings of the IEEE/CVF international conference on computer vision},
  pages={2085--2094},
  year={2021}
}

@article{ho2020denoising,
  title={Denoising diffusion probabilistic models},
  author={Ho, Jonathan and Jain, Ajay and Abbeel, Pieter},
  journal={Advances in neural information processing systems},
  volume={33},
  pages={6840--6851},
  year={2020}
}

@inproceedings{rombach2022high,
  title={High-resolution image synthesis with latent diffusion models},
  author={Rombach, Robin and Blattmann, Andreas and Lorenz, Dominik and Esser, Patrick and Ommer, Bj{\"o}rn},
  booktitle={Proceedings of the IEEE/CVF conference on computer vision and pattern recognition},
  pages={10684--10695},
  year={2022}
}

@inproceedings{lugmayr2022repaint,
  title={Repaint: Inpainting using denoising diffusion probabilistic models},
  author={Lugmayr, Andreas and Danelljan, Martin and Romero, Andres and Yu, Fisher and Timofte, Radu and Van Gool, Luc},
  booktitle={Proceedings of the IEEE/CVF conference on computer vision and pattern recognition},
  pages={11461--11471},
  year={2022}
}

@article{meng2021sdedit,
  title={Sdedit: Guided image synthesis and editing with stochastic differential equations},
  author={Meng, Chenlin and He, Yutong and Song, Yang and Song, Jiaming and Wu, Jiajun and Zhu, Jun-Yan and Ermon, Stefano},
  journal={arXiv preprint arXiv:2108.01073},
  year={2021}
}

@article{telea2004image,
  title={An image inpainting technique based on the fast marching method},
  author={Telea, Alexandru},
  journal={Journal of graphics tools},
  volume={9},
  number={1},
  pages={23--34},
  year={2004},
  publisher={Taylor \& Francis}
}

@article{ho2022classifier,
  title={Classifier-free diffusion guidance},
  author={Ho, Jonathan and Salimans, Tim},
  journal={arXiv preprint arXiv:2207.12598},
  year={2022}
}

@article{qin2020u2,
  title={U2-Net: Going deeper with nested U-structure for salient object detection},
  author={Qin, Xuebin and Zhang, Zichen and Huang, Chenyang and Dehghan, Masood and Zaiane, Osmar R and Jagersand, Martin},
  journal={Pattern recognition},
  volume={106},
  pages={107404},
  year={2020},
  publisher={Elsevier}
}

@article{fischler1981random,
  title={Random sample consensus: a paradigm for model fitting with applications to image analysis and automated cartography},
  author={Fischler, Martin A and Bolles, Robert C},
  journal={Communications of the ACM},
  volume={24},
  number={6},
  pages={381--395},
  year={1981},
  publisher={ACM New York, NY, USA}
}

@article{gemini,
  title={Gemini: a family of highly capable multimodal models},
  author={Team, Gemini and Anil, Rohan and Borgeaud, Sebastian and Alayrac, Jean-Baptiste and Yu, Jiahui and Soricut, Radu and Schalkwyk, Johan and Dai, Andrew M and Hauth, Anja and Millican, Katie and others},
  journal={arXiv preprint arXiv:2312.11805},
  year={2023}
}

@inproceedings{brooks2023instructpix2pix,
  title={Instructpix2pix: Learning to follow image editing instructions},
  author={Brooks, Tim and Holynski, Aleksander and Efros, Alexei A},
  booktitle={Proceedings of the IEEE/CVF conference on computer vision and pattern recognition},
  pages={18392--18402},
  year={2023}
}

@inproceedings{zhang2023controlnet,
  title={Adding conditional control to text-to-image diffusion models},
  author={Zhang, Lvmin and Rao, Anyi and Agrawala, Maneesh},
  booktitle={Proceedings of the IEEE/CVF international conference on computer vision},
  pages={3836--3847},
  year={2023}
}

@inproceedings{ranftl2021vision,
  title={Vision transformers for dense prediction},
  author={Ranftl, Ren{\'e} and Bochkovskiy, Alexey and Koltun, Vladlen},
  booktitle={Proceedings of the IEEE/CVF international conference on computer vision},
  pages={12179--12188},
  year={2021}
}

@inproceedings{pan2023drag,
  title={Drag your gan: Interactive point-based manipulation on the generative image manifold},
  author={Pan, Xingang and Tewari, Ayush and Leimk{\"u}hler, Thomas and Liu, Lingjie and Meka, Abhimitra and Theobalt, Christian},
  booktitle={ACM SIGGRAPH 2023 conference proceedings},
  pages={1--11},
  year={2023}
}

@inproceedings{shi2024dragdiffusion,
  title={Dragdiffusion: Harnessing diffusion models for interactive point-based image editing},
  author={Shi, Yujun and Xue, Chuhui and Liew, Jun Hao and Pan, Jiachun and Yan, Hanshu and Zhang, Wenqing and Tan, Vincent YF and Bai, Song},
  booktitle={Proceedings of the IEEE/CVF conference on computer vision and pattern recognition},
  pages={8839--8849},
  year={2024}
}

@article{couairon2022diffedit,
  title={Diffedit: Diffusion-based semantic image editing with mask guidance},
  author={Couairon, Guillaume and Verbeek, Jakob and Schwenk, Holger and Cord, Matthieu},
  journal={arXiv preprint arXiv:2210.11427},
  year={2022}
}

@article{michel2023object,
  title={Object 3dit: Language-guided 3d-aware image editing},
  author={Michel, Oscar and Bhattad, Anand and VanderBilt, Eli and Krishna, Ranjay and Kembhavi, Aniruddha and Gupta, Tanmay},
  journal={Advances in Neural Information Processing Systems},
  volume={36},
  pages={3497--3516},
  year={2023}
}

@article{chang2015shapenet,
  title={Shapenet: An information-rich 3d model repository},
  author={Chang, Angel X and Funkhouser, Thomas and Guibas, Leonidas and Hanrahan, Pat and Huang, Qixing and Li, Zimo and Savarese, Silvio and Savva, Manolis and Song, Shuran and Su, Hao and others},
  journal={arXiv preprint arXiv:1512.03012},
  year={2015}
}

@misc{liu2023zero1to3zeroshotimage3d,
      title={Zero-1-to-3: Zero-shot One Image to 3D Object}, 
      author={Ruoshi Liu and Rundi Wu and Basile Van Hoorick and Pavel Tokmakov and Sergey Zakharov and Carl Vondrick},
      year={2023},
      eprint={2303.11328},
      archivePrefix={arXiv},
      primaryClass={cs.CV},
      url={https://arxiv.org/abs/2303.11328}, 
}

@inproceedings{shen2020interpreting,
  title={Interpreting the latent space of gans for semantic face editing},
  author={Shen, Yujun and Gu, Jinjin and Tang, Xiaoou and Zhou, Bolei},
  booktitle={Proceedings of the IEEE/CVF conference on computer vision and pattern recognition},
  pages={9243--9252},
  year={2020}
}

@misc{xie2015holisticallynestededgedetection,
      title={Holistically-Nested Edge Detection}, 
      author={Saining Xie and Zhuowen Tu},
      year={2015},
      eprint={1504.06375},
      archivePrefix={arXiv},
      primaryClass={cs.CV},
      url={https://arxiv.org/abs/1504.06375}, 
}

@article{liu2023syncdreamer,
  title={Syncdreamer: Generating multiview-consistent images from a single-view image},
  author={Liu, Yuan and Lin, Cheng and Zeng, Zijiao and Long, Xiaoxiao and Liu, Lingjie and Komura, Taku and Wang, Wenping},
  journal={arXiv preprint arXiv:2309.03453},
  year={2023}
}

@inproceedings{sajnani2025geodiffuser,
  title={Geodiffuser: Geometry-based image editing with diffusion models},
  author={Sajnani, Rahul and Vanbaar, Jeroen and Min, Jie and Katyal, Kapil D and Sridhar, Srinath},
  booktitle={Proceedings of the Winter Conference on Applications of Computer Vision},
  pages={472--482},
  year={2025}
}

@inproceedings{pandey2024diffusion,
  title={Diffusion handles enabling 3d edits for diffusion models by lifting activations to 3d},
  author={Pandey, Karran and Guerrero, Paul and Gadelha, Matheus and Hold-Geoffroy, Yannick and Singh, Karan and Mitra, Niloy J},
  booktitle={Proceedings of the IEEE/CVF Conference on Computer Vision and Pattern Recognition},
  pages={7695--7704},
  year={2024}
}

@article{chen2025blenderfusion,
  title={Blenderfusion: 3d-grounded visual editing and generative compositing},
  author={Chen, Jiacheng and Mehran, Ramin and Jia, Xuhui and Xie, Saining and Woo, Sanghyun},
  journal={arXiv preprint arXiv:2506.17450},
  year={2025}
}

@inproceedings{yu2025objectmover,
  title={Objectmover: Generative object movement with video prior},
  author={Yu, Xin and Wang, Tianyu and Kim, Soo Ye and Guerrero, Paul and Chen, Xi and Liu, Qing and Lin, Zhe and Qi, Xiaojuan},
  booktitle={Proceedings of the IEEE/CVF Conference on Computer Vision and Pattern Recognition},
  pages={17682--17691},
  year={2025}
}

@article{ravi2024sam,
  title={Sam 2: Segment anything in images and videos},
  author={Ravi, Nikhila and Gabeur, Valentin and Hu, Yuan-Ting and Hu, Ronghang and Ryali, Chaitanya and Ma, Tengyu and Khedr, Haitham and R{\"a}dle, Roman and Rolland, Chloe and Gustafson, Laura and others},
  journal={arXiv preprint arXiv:2408.00714},
  year={2024}
}

@inproceedings{deitke2023objaverse,
  title={Objaverse: A universe of annotated 3d objects},
  author={Deitke, Matt and Schwenk, Dustin and Salvador, Jordi and Weihs, Luca and Michel, Oscar and VanderBilt, Eli and Schmidt, Ludwig and Ehsani, Kiana and Kembhavi, Aniruddha and Farhadi, Ali},
  booktitle={Proceedings of the IEEE/CVF conference on computer vision and pattern recognition},
  pages={13142--13153},
  year={2023}
}

@inproceedings{wang2025vggt,
  title={Vggt: Visual geometry grounded transformer},
  author={Wang, Jianyuan and Chen, Minghao and Karaev, Nikita and Vedaldi, Andrea and Rupprecht, Christian and Novotny, David},
  booktitle={Proceedings of the Computer Vision and Pattern Recognition Conference},
  pages={5294--5306},
  year={2025}
}

@inproceedings{voleti2024sv3d,
  title={Sv3d: Novel multi-view synthesis and 3d generation from a single image using latent video diffusion},
  author={Voleti, Vikram and Yao, Chun-Han and Boss, Mark and Letts, Adam and Pankratz, David and Tochilkin, Dmitry and Laforte, Christian and Rombach, Robin and Jampani, Varun},
  booktitle={European Conference on Computer Vision},
  pages={439--457},
  year={2024},
  organization={Springer}
}

@article{cao2026freeorbit4d,
  title={FreeOrbit4D: Training-Free Arbitrary Camera Redirection for Monocular Videos via Geometry-Complete 4D Reconstruction},
  author={Cao, Wei and Zhang, Hao and Tian, Fengrui and Wu, Yulun and Li, Yingying and Wang, Shenlong and Yu, Ning and Liu, Yaoyao},
  journal={arXiv preprint arXiv:2601.18993},
  year={2026}
}

@article{wan2025wan,
  title={Wan: Open and advanced large-scale video generative models},
  author={Wan, Team and Wang, Ang and Ai, Baole and Wen, Bin and Mao, Chaojie and Xie, Chen-Wei and Chen, Di and Yu, Feiwu and Zhao, Haiming and Yang, Jianxiao and others},
  journal={arXiv preprint arXiv:2503.20314},
  year={2025}
}

@inproceedings{jiang2025vace,
  title={Vace: All-in-one video creation and editing},
  author={Jiang, Zeyinzi and Han, Zhen and Mao, Chaojie and Zhang, Jingfeng and Pan, Yulin and Liu, Yu},
  booktitle={Proceedings of the IEEE/CVF International Conference on Computer Vision},
  pages={17191--17202},
  year={2025}
}

@inproceedings{avrahami2022blended,
  title={Blended diffusion for text-driven editing of natural images},
  author={Avrahami, Omri and Lischinski, Dani and Fried, Ohad},
  booktitle={Proceedings of the IEEE/CVF conference on computer vision and pattern recognition},
  pages={18208--18218},
  year={2022}
}

@inproceedings{yang2023contranerf,
  title={Contranerf: Generalizable neural radiance fields for synthetic-to-real novel view synthesis via contrastive learning},
  author={Yang, Hao and Hong, Lanqing and Li, Aoxue and Hu, Tianyang and Li, Zhenguo and Lee, Gim Hee and Wang, Liwei},
  booktitle={Proceedings of the IEEE/CVF conference on computer vision and pattern recognition},
  pages={16508--16517},
  year={2023}
}

@inproceedings{wiles2020synsin,
  title={Synsin: End-to-end view synthesis from a single image},
  author={Wiles, Olivia and Gkioxari, Georgia and Szeliski, Richard and Johnson, Justin},
  booktitle={Proceedings of the IEEE/CVF conference on computer vision and pattern recognition},
  pages={7467--7477},
  year={2020}
}

@article{singer2025time,
  title={Time-to-Move: Training-Free Motion Controlled Video Generation via Dual-Clock Denoising},
  author={Singer, Assaf and Rotstein, Noam and Mann, Amir and Kimmel, Ron and Litany, Or},
  journal={arXiv preprint arXiv:2511.08633},
  year={2025}
}

@inproceedings{peebles2023dit,
  title={Scalable Diffusion Models with Transformers},
  author={Peebles, William and Xie, Saining},
  booktitle={Proceedings of the IEEE/CVF International Conference on Computer Vision},
  pages={4195--4205},
  year={2023}
}

@inproceedings{caron2021dino,
  title={Emerging properties in self-supervised vision transformers},
  author={Caron, Mathilde and Touvron, Hugo and Misra, Ishan and J{\'e}gou, Herv{\'e} and Mairal, Julien and Bojanowski, Piotr and Joulin, Armand},
  booktitle={Proceedings of the IEEE/CVF international conference on computer vision},
  pages={9650--9660},
  year={2021}
}

@inproceedings{radford2021clip,
  title={Learning transferable visual models from natural language supervision},
  author={Radford, Alec and Kim, Jong Wook and Hallacy, Chris and Ramesh, Aditya and Goh, Gabriel and Agarwal, Sandhini and Sastry, Girish and Askell, Amanda and Mishkin, Pamela and Clark, Jack and others},
  booktitle={International conference on machine learning},
  pages={8748--8763},
  year={2021},
  organization={PmLR}
}

@article{fu2023dreamsim,
  title={Dreamsim: Learning new dimensions of human visual similarity using synthetic data},
  author={Fu, Stephanie and Tamir, Netanel and Sundaram, Shobhita and Chai, Lucy and Zhang, Richard and Dekel, Tali and Isola, Phillip},
  journal={arXiv preprint arXiv:2306.09344},
  year={2023}
}

@inproceedings{yenphraphai2024image,
  title={Image sculpting: Precise object editing with 3d geometry control},
  author={Yenphraphai, Jiraphon and Pan, Xichen and Liu, Sainan and Panozzo, Daniele and Xie, Saining},
  booktitle={Proceedings of the IEEE/CVF Conference on Computer Vision and Pattern Recognition},
  pages={4241--4251},
  year={2024}
}

@article{labs2025flux,
  title={Flux. 1 kontext: Flow matching for in-context image generation and editing in latent space},
  author={Batifol, Stephen and Blattmann, Andreas and Boesel, Frederic and Consul, Saksham and Diagne, Cyril and Dockhorn, Tim and English, Jack and English, Zion and Esser, Patrick and Kulal, Sumith and others},
  journal={arXiv e-prints},
  pages={arXiv--2506},
  year={2025}
}

@article{wu2025qwenimagetechnicalreport,
  title={Qwen-image technical report},
  author={Wu, Chenfei and Li, Jiahao and Zhou, Jingren and Lin, Junyang and Gao, Kaiyuan and Yan, Kun and Yin, Sheng-ming and Bai, Shuai and Xu, Xiao and Chen, Yilei and others},
  journal={arXiv preprint arXiv:2508.02324},
  year={2025}
}

@misc{nanobanana,
  title        = {Gemini 2.5 Flash Image (Nano Banana) | Google AI Studio},
  howpublished = {\url{https://aistudio.google.com/models/gemini-2-5-flash-image}}
}

@article{ctrl_shift,
  title={Ctrl\&Shift: High-Quality Geometry-Aware Object Manipulation in Visual Generation},
  author={Ruan, Penghui and Zi, Bojia and Qi, Xianbiao and Huang, Youze and Xiao, Rong and Wang, Pichao and Cao, Jiannong and Shi, Yuhui},
  journal={arXiv preprint arXiv:2602.11440},
  year={2026}
}

@inproceedings{fatezero,
  title={Fatezero: Fusing attentions for zero-shot text-based video editing},
  author={Qi, Chenyang and Cun, Xiaodong and Zhang, Yong and Lei, Chenyang and Wang, Xintao and Shan, Ying and Chen, Qifeng},
  booktitle={Proceedings of the IEEE/CVF International Conference on Computer Vision},
  pages={15932--15942},
  year={2023}
}

@inproceedings{3ditscene,
  title={3ditscene: Editing any scene via language-guided disentangled gaussian splatting},
  author={Zhang, Qihang and Xu, Yinghao and Wang, Chaoyang and Lee, Hsin-Ying and Wetzstein, Gordon and Zhou, Bolei and Yang, Ceyuan},
  booktitle={International Conference on Learning Representations},
  volume={2025},
  pages={2760--2775},
  year={2025}
}

@article{team2023gemini,
  title={Gemini: a family of highly capable multimodal models},
  author={Team, Gemini and Anil, Rohan and Borgeaud, Sebastian and Alayrac, Jean-Baptiste and Yu, Jiahui and Soricut, Radu and Schalkwyk, Johan and Dai, Andrew M and Hauth, Anja and Millican, Katie and others},
  journal={arXiv preprint arXiv:2312.11805},
  year={2023}
}

@article{zhang2025objectadd,
  title={Objectadd: adding objects into image via a training-free diffusion modification fashion},
  author={Zhang, Ziyue and Lin, Mingbao and Song, Quanjian and Zhang, Yuxin and Ji, Rongrong},
  journal={Pattern Recognition},
  pages={112807},
  year={2025},
  publisher={Elsevier}
}

@article{ma2026group,
  title={Group Editing: Edit Multiple Images in One Go},
  author={Ma, Yue and Wang, Xinyu and Ma, Qianli and Wang, Qinghe and Zheng, Mingzhe and Yang, Xiangpeng and Li, Hao and Zhao, Chongbo and Ying, Jixuan and Yang, Harry and others},
  journal={arXiv preprint arXiv:2603.22883},
  year={2026}
}

@inproceedings{ma2024followpose,
  title={Follow your pose: Pose-guided text-to-video generation using pose-free videos},
  author={Ma, Yue and He, Yingqing and Cun, Xiaodong and Wang, Xintao and Chen, Siran and Li, Xiu and Chen, Qifeng},
  booktitle={Proceedings of the AAAI Conference on Artificial Intelligence},
  volume={38},
  number={5},
  pages={4117--4125},
  year={2024}
}

@article{ma2025followcreation,
  title={Follow-Your-Creation: Empowering 4D Creation through Video Inpainting},
  author={Ma, Yue and Feng, Kunyu and Zhang, Xinhua and Liu, Hongyu and Zhang, David Junhao and Xing, Jinbo and Zhang, Yinhan and Yang, Ayden and Wang, Zeyu and Chen, Qifeng},
  journal={arXiv preprint arXiv:2506.04590},
  year={2025}
}

@article{ma2026fastvmt,
  title={FastVMT: Eliminating Redundancy in Video Motion Transfer},
  author={Ma, Yue and Wang, Zhikai and Ren, Tianhao and Zheng, Mingzhe and Liu, Hongyu and Guo, Jiayi and Fong, Mark and Xue, Yuxuan and Zhao, Zixiang and Schindler, Konrad and others},
  journal={arXiv preprint arXiv:2602.05551},
  year={2026}
}

@article{ma2025followyourmotion,
  title={Follow-Your-Motion: Video Motion Transfer via Efficient Spatial-Temporal Decoupled Finetuning},
  author={Ma, Yue and Liu, Yulong and Zhu, Qiyuan and Yang, Ayden and Feng, Kunyu and Zhang, Xinhua and Li, Zhifeng and Han, Sirui and Qi, Chenyang and Chen, Qifeng},
  journal={arXiv preprint arXiv:2506.05207},
  year={2025}
}

@article{ma2025controllable,
  title={Controllable Video Generation: A Survey},
  author={Ma, Yue and Feng, Kunyu and Hu, Zhongyuan and Wang, Xinyu and Wang, Yucheng and Zheng, Mingzhe and He, Xuanhua and Zhu, Chenyang and Liu, Hongyu and He, Yingqing and others},
  journal={arXiv preprint arXiv:2507.16869},
  year={2025}
}

@article{ma2025followfaster,
  title={Follow-your-emoji-faster: Towards efficient, fine-controllable, and expressive freestyle portrait animation},
  author={Ma, Yue and Yan, Zexuan and Liu, Hongyu and Wang, Hongfa and Pan, Heng and He, Yingqing and Yuan, Junkun and Zeng, Ailing and Cai, Chengfei and Shum, Heung-Yeung and others},
  journal={arXiv preprint arXiv:2509.16630},
  year={2025}
}

@inproceedings{ma2024followyouremoji,
  title={Follow-your-emoji: Fine-controllable and expressive freestyle portrait animation},
  author={Ma, Yue and Liu, Hongyu and Wang, Hongfa and Pan, Heng and He, Yingqing and Yuan, Junkun and Zeng, Ailing and Cai, Chengfei and Shum, Heung-Yeung and Liu, Wei and others},
  booktitle={SIGGRAPH Asia 2024 Conference Papers},
  pages={1--12},
  year={2024}
}

@inproceedings{ma2025followyourclick,
  title={Follow-Your-Click: Open-domain Regional Image Animation via Motion Prompts},
  author={Ma, Yue and He, Yingqing and Wang, Hongfa and Wang, Andong and Shen, Leqi and Qi, Chenyang and Ying, Jixuan and Cai, Chengfei and Li, Zhifeng and Shum, Heung-Yeung and others},
  booktitle={Proceedings of the AAAI Conference on Artificial Intelligence},
  volume={39},
  number={6},
  pages={6018--6026},
  year={2025}
}

@article{shen2025follow,
  title={Follow-Your-Preference: Towards Preference-Aligned Image Inpainting},
  author={Shen, Yutao and Yuan, Junkun and Aonishi, Toru and Nakayama, Hideki and Ma, Yue},
  journal={arXiv preprint arXiv:2509.23082},
  year={2025}
}

@article{chen2025contextflow,
  title={ContextFlow: Training-Free Video Object Editing via Adaptive Context Enrichment},
  author={Chen, Yiyang and He, Xuanhua and Ma, Xiujun and Ma, Yue},
  journal={arXiv preprint arXiv:2509.17818},
  year={2025}
}

@article{he2024id,
  title={Id-animator: Zero-shot identity-preserving human video generation},
  author={He, Xuanhua and Liu, Quande and Qian, Shengju and Wang, Xin and Hu, Tao and Cao, Ke and Yan, Keyu and Zhang, Jie},
  journal={arXiv preprint arXiv:2404.15275},
  year={2024}
}

@article{he2025fulldit2,
  title={Fulldit2: Efficient in-context conditioning for video diffusion transformers},
  author={He, Xuanhua and Liu, Quande and Ye, Zixuan and Ye, Weicai and Wang, Qiulin and Wang, Xintao and Chen, Qifeng and Wan, Pengfei and Zhang, Di and Gai, Kun},
  journal={arXiv preprint arXiv:2506.04213},
  year={2025}
}

@article{ye2025unic,
  title={Unic: Unified in-context video editing},
  author={Ye, Zixuan and He, Xuanhua and Liu, Quande and Wang, Qiulin and Wang, Xintao and Wan, Pengfei and Zhang, Di and Gai, Kun and Chen, Qifeng and Luo, Wenhan},
  journal={ICLR 2026},
  year={2025}
}

@article{song2026vista,
  title={VISTA: Triplet-Supervised Video Style Transfer with Diffusion Transformers},
  author={Song, Yiren and Yao, Wangzi and Wang, Haofan and Shou, Mike Zheng},
  journal={arXiv preprint arXiv:2605.17312},
  year={2026}
}

@article{zhao2026dydit,
  title={DyDiT++: Diffusion Transformers with Timestep and Spatial Dynamics for Efficient Visual Generation},
  author={Zhao, Wangbo and Han, Yizeng and Tang, Jiasheng and Wang, Kai and Luo, Hao and Song, Yibing and Huang, Gao and Wang, Fan and You, Yang},
  journal={TPAMI},
  year={2026},
  publisher={IEEE}
}

@inproceedings{dynamictuning_rapid,
  title={RAPID\^{} 3: Tri-Level Reinforced Acceleration Policies for Diffusion Transformer},
  author={Zhao, Wangbo and Han, Yizeng and Tang, Zhiwei and Tang, Jiasheng and others},
  booktitle={ICLR},
  year={2026}
}

@inproceedings{nan2026accelerating,
  title={Accelerating Autoregressive Video Diffusion via History-Guided Cache and Residual Correction},
  author={Nan, Kepan and Zhao, Wangbo and Zhou, Penghao and Li, Jun and Yang, Zhenheng and Yang, Jian and Tai, Ying},
  booktitle={CVPR},
  pages={43740--43750},
  year={2026}
}

@article{song2026streamingeffect,
  title={StreamingEffect: Real-Time Human-Centric Video Effect Generation},
  author={Song, Yiren and Liu, Cheng and Jiang, Yuxin and Shou, Mike Zheng},
  journal={arXiv preprint arXiv:2605.17019},
  year={2026}
}

@article{gao2026pai,
  title={PAI-Studio: Cinematic Video Background Replacement with Camera-Aware Motion},
  author={Gao, Heyuan and Tang, Bangxun and Song, Yiren and Fang, Guian and He, Zijian and Yang, Jie and Shou, Mike Zheng},
  journal={arXiv preprint arXiv:2606.01399},
  year={2026}
}

@inproceedings{song2024processpainter,
  title={ProcessPainter: Learning to draw from sequence data},
  author={Song, Yiren and Huang, Shijie and Yao, Chen and Ci, Hai and Ye, Xiaojun and Liu, Jiaming and Zhang, Yuxuan and Shou, Mike Zheng},
  booktitle={SIGGRAPH Asia 2024 Conference Papers},
  pages={1--10},
  year={2024}
}

@article{wang2024taming,
  title={Taming rectified flow for inversion and editing},
  author={Wang, Jiangshan and Pu, Junfu and Qi, Zhongang and Guo, Jiayi and Ma, Yue and Huang, Nisha and Chen, Yuxin and Li, Xiu and Shan, Ying},
  journal={arXiv preprint arXiv:2411.04746},
  year={2024}
}

@article{li2025lodge,
  title={Lodge++: High-Quality and Long Dance Generation With Robust Choreography Patterns},
  author={Li, Ronghui and Zhang, Hongwen and Zhang, Yachao and Zhang, Yuxiang and Zhang, Youliang and Guo, Jie and Zhang, Yan and Li, Xiu and Liu, Yebin},
  journal={IEEE Transactions on Pattern Analysis and Machine Intelligence},
  year={2025},
  publisher={IEEE}
}

@inproceedings{li2024lodge,
  title={Lodge: A coarse to fine diffusion network for long dance generation guided by the characteristic dance primitives},
  author={Li, Ronghui and Zhang, YuXiang and Zhang, Yachao and Zhang, Hongwen and Guo, Jie and Zhang, Yan and Liu, Yebin and Li, Xiu},
  booktitle={Proceedings of the IEEE/CVF Conference on Computer Vision and Pattern Recognition},
  pages={1524--1534},
  year={2024}
}

@inproceedings{li2023finedance,
  title={Finedance: A fine-grained choreography dataset for 3d full body dance generation},
  author={Li, Ronghui and Zhao, Junfan and Zhang, Yachao and Su, Mingyang and Ren, Zeping and Zhang, Han and Tang, Yansong and Li, Xiu},
  booktitle={Proceedings of the IEEE/CVF International Conference on Computer Vision},
  pages={10234--10243},
  year={2023}
}

@inproceedings{xu2026smrabooth,
  title={Smrabooth: Subject and motion representation alignment for customized video generation},
  author={Xu, Xuancheng and Li, Yaning and You, Sisi and Bao, Bing-Kun},
  booktitle={Proceedings of the IEEE/CVF Conference on Computer Vision and Pattern Recognition},
  pages={16130--16141},
  year={2026}
}

@misc{huang2026exposurebiasalleviatedirectional,
      title={Exposure Bias Can Alleviate Itself via Directional and Frequency Rectification in Flow Matching}, 
      author={Guanbo Huang and Jingjia Mao and Fanding Huang and Fengkai Liu and Xiangyang Luo and Yaoyuan Liang and Jiasheng Lu and Xiaoe Wang and Pei Liu and Ruiliu Fu and Ruqi Huang and Shao-Lun Huang},
      year={2026},
      eprint={2606.28226},
      archivePrefix={arXiv},
      primaryClass={cs.CV},
      url={https://arxiv.org/abs/2606.28226}, 
}

@inproceedings{feng2025dit4edit,
  title={Dit4edit: Diffusion transformer for image editing},
  author={Feng, Kunyu and Ma, Yue and Wang, Bingyuan and Qi, Chenyang and Chen, Haozhe and Chen, Qifeng and Wang, Zeyu},
  booktitle={Proceedings of the AAAI Conference on Artificial Intelligence},
  volume={39},
  number={3},
  pages={2969--2977},
  year={2025}
}

@article{wang2024cove,
  title={Cove: Unleashing the diffusion feature correspondence for consistent video editing},
  author={Wang, Jiangshan and Ma, Yue and Guo, Jiayi and Xiao, Yicheng and Huang, Gao and Li, Xiu},
  journal={Advances in Neural Information Processing Systems},
  volume={37},
  pages={96541--96565},
  year={2024}
}

@inproceedings{ma2025magicstick,
  title={Magicstick: Controllable video editing via control handle transformations},
  author={Ma, Yue and Cun, Xiaodong and Liang, Sen and Xing, Jinbo and He, Yingqing and Qi, Chenyang and Chen, Siran and Chen, Qifeng},
  booktitle={2025 IEEE/CVF Winter Conference on Applications of Computer Vision (WACV)},
  pages={9385--9395},
  year={2025},
  organization={IEEE}
}

@article{yang2025unified,
  title={Unified Video Editing with Temporal Reasoner},
  author={Yang, Xiangpeng and Xie, Ji and Yang, Yiyuan and Huang, Yan and Xu, Min and Wu, Qiang},
  journal={arXiv preprint arXiv:2512.07469},
  year={2025}
}
}


\clearpage
\appendix 
\section*{Appendix}

\setcounter{figure}{0}   
\setcounter{table}{0}    
\setcounter{equation}{0} 

\renewcommand{\thefigure}{A.\arabic{figure}}
\renewcommand{\thetable}{A.\arabic{table}}
\renewcommand{\theequation}{A.\arabic{equation}}
\section{Extended Implementation Details}
\label{sec:supp_implementation}

To ensure full reproducibility of our \textbf{GeoEdit} framework, we provide comprehensive algorithmic settings, hyper-parameters, and computational profiles that extend the brief descriptions in the main text.

\subsection{Object Masking and Background Inpainting}
For target object isolation, we dynamically select the segmentation strategy based on scene complexity. For single-object scenes, we utilize the off-the-shelf \texttt{rembg}\footnote{\url{https://github.com/danielgatis/rembg}} tool, which is powered by U$^2$-Net~\cite{qin2020u2}. For scenarios involving complex or multiple objects, we employ the Segment Anything Model 2 (SAM2)~\cite{ravi2024sam} to obtain precise binary masks. When rendering the geometric proxy, the disoccluded background hole is filled using the Telea inpainting algorithm~\cite{telea2004image}. To preserve local structural continuity without introducing excessive blurring, the inpainting neighborhood radius is strictly set to $3$ pixels.

\subsection{Monocular Lifting and Multi-View Synthesis}
\noindent\textbf{VGGT for Scene Lifting.} The VGGT~\cite{wang2025vggt} backbone scales and pads input images to a resolution of $518 \times 518$ before inference. Rather than assuming a fixed Field of View (FOV), VGGT dynamically predicts the focal lengths ($f_x, f_y$) from the encoded pose representations. For structural stability, the camera intrinsics are constrained with a zero-skew assumption, and the principal point ($c_x, c_y$) is fixed at the exact image center $(W/2, H/2)$. To suppress noisy point cloud artifacts, we filter the lifted geometry using mask erosion combined with a strict depth confidence threshold.

\noindent\textbf{SV3D for Object Completion.} We adopt the \texttt{sv3d\_u} variant of SV3D~\cite{voleti2024sv3d} for canonical object completion. The model generates a fixed sequence of $21$ novel views over $50$ inference steps. Notably, the Classifier-Free Guidance (CFG)~\cite{ho2022classifier} scale is not a static constant; we utilize a \texttt{TrianglePredictionGuider} that dynamically modulates the scale curve from $1.0$ to $2.5$ across the generated frames.

\noindent\textbf{Correspondence and Alignment.} Instead of standard RANSAC~\cite{fischler1981random}, our 3D precise alignment computes the optimal similarity transformation, including rotation, translation, and scale, through deterministic least-squares optimization on the matched correspondences. To robustly reject outliers, we calculate the spatial residuals and apply a strict percentile threshold, retaining only the top 95\% of the most accurate inliers.

\subsection{Dual-Branch Denoising via Video Diffusion Prior}
The terminal denoising phase is driven by a state-of-the-art large-scale video diffusion model~\cite{wan2025wan, jiang2025vace}. To faithfully enforce the rigid 3D skeleton of the edited object, we empirically found that the temporal prior of the video diffusion model is strictly necessary. We force the model to generate a sequence of 81 frames (matching its default training horizon), as generating a drastically reduced frame count (\eg, 1 frame) completely collapses the structural adherence. We then extract the last frame of this sequence as our final image composite.

The generation is steered by descriptive text prompts and refined by comprehensive negative prompts. These negative constraints are crucial for suppressing common diffusion artifacts, specifically targeting structural mutations (\eg, fused fingers, malformed limbs), poor rendering quality (\eg, JPEG artifacts, overexposure), and undesired temporal or stylistic shifts (\eg, flickering, cluttered backgrounds, text/subtitles). 

For the sampling configuration, we utilize a native \texttt{FlowMatchScheduler} corresponding to the employed video backbone. The reverse process spans 50 steps with a constant CFG scale of 5.0. As established in the main text, our variance-homogeneous injection strictly operates within the interval $(t_{weak}=47, t_{strong}=40)$.

\subsection{Computational Cost and Resource Profiling}
Our fully training-free pipeline evaluates efficiently on a single NVIDIA A800 GPU. The computational profile for generating a standardized $720 \times 480$ composite is bifurcated as follows:
\begin{itemize}
    \item \textbf{Lifting and Completion (VGGT + SV3D):} The 3D scene reconstruction and 21-frame multi-view synthesis require approximately 19 GB of VRAM and are completed in $\sim$2 minutes.
    \item \textbf{Dual-Branch Denoising (Video Backbone):} The 81-frame generation using the large-scale video diffusion model requires approximately 44 GB of VRAM, with an inference latency of $\sim$16 minutes.
\end{itemize}
While the video backbone introduces a temporal computational overhead, it entirely eliminates the need for test-time fine-tuning or instance-specific optimization, achieving zero-shot geometry-aware editing.

\subsection{Quantifying Attention Leakage (ALR)}
In the main text, we claim that the variance-homogeneous injection mitigates self-attention leakage. To rigorously quantify this, we defined the Attention Leakage Ratio (ALR). Let $A^{l,h} \in \mathbb{R}^{N \times N}$ be the spatial self-attention map at layer $l$ and head $h$. We denote $\mathcal{B}$ as the set of background tokens and $\mathcal{F}$ as the set of foreground proxy tokens. ALR calculates the fraction of attention mass that background queries undesirably assign to foreground tokens:
\begin{equation}
    ALR = \operatorname{mean}_{l,h} \left( \frac{\sum_{i \in \mathcal{B}, j \in \mathcal{F}} A_{i,j}^{l,h}}{\sum_{i \in \mathcal{B}, j \in *} A_{i,j}^{l,h}} \right)
\end{equation}
Our empirical analysis demonstrates that without variance-matching, the ALR surges significantly, causing severe background blur. Conversely, our synchronized injection reduces the ALR from $8.4\%$ to $6.4\%$ at the peak leakage step ($t=46$), verifying its efficacy in preserving distinct foreground-background generative trajectories.

\section{Benchmark Design}
\begin{figure*}[t]
  \centering
  \makebox[\textwidth][c]{%
    \includegraphics[
      width=1.15\textwidth,
      trim=4.5cm 0cm 4.5cm 0cm,
      clip
    ]{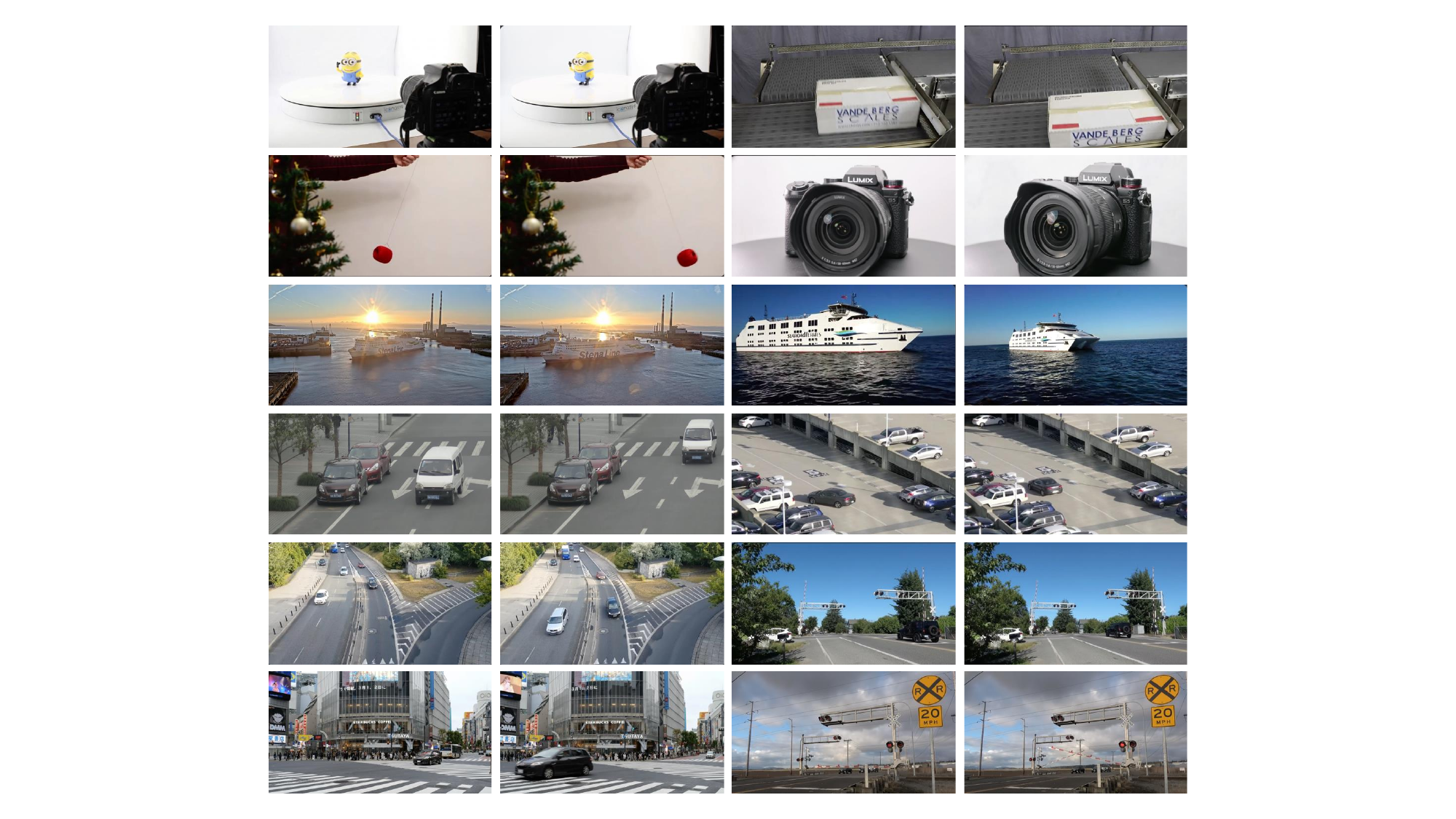}
  }
  \caption{\textbf{GeoEditBench.} GeoEditBench consists of 200 curated image pairs collected from the web, with prominent geometric cues, rich real-world textures, and diverse lighting conditions.}
  \label{fig:benchmark}
\end{figure*}

\noindent\textbf{Additional benchmark visualization.}
To complement the description in the main paper, we provide additional visual examples from GeoEditBench in Fig.~\ref{fig:benchmark}. 
The benchmark consists of 200 carefully curated image pairs collected from the web, featuring prominent geometric cues, rich real-world textures, and diverse lighting conditions. 
The dataset spans three representative spatial editing scenarios, including object translation, object rotation, and camera movement, enabling evaluation across a wide range of geometric transformations in realistic scenes.

\section{Comparison with Baselines}
\label{sec:comparison}

We strictly follow the official configurations of all compared baselines (\eg, Flux-Kontext~\cite{labs2025flux}, Qwen-Image~\cite{wu2025qwenimagetechnicalreport}, NanoBanana~\cite{nanobanana}, 3DiT~\cite{michel2023object}, and Image-Sculpting~\cite{yenphraphai2024image}) and use default hyperparameters whenever possible. For methods requiring tuning, we apply a limited grid search within recommended ranges. Prompt designs are standardized across all methods to ensure consistent input conditions.

\vspace{2mm}
\noindent\textbf{Evaluation on External 3DEdit-Bench.} 
To demonstrate the generalization ability of our framework, we further evaluate GeoEdit on a stratified 30-case subset from the external 3DEdit-Bench. As reported in Tab.~\ref{tab:3dedit_bench}, GeoEdit consistently achieves state-of-the-art performance across geometric and perceptual metrics when compared to the highly competitive 3DitScene (which relies on per-scene SDS optimization) and NanoBanana. Notably, our training-free approach significantly outperforms 3DitScene in 3D physical correctness metrics (Object IoU) while requiring a fraction of the computational time.

\begin{figure}[t]
  \centering
  \includegraphics[width=\linewidth]{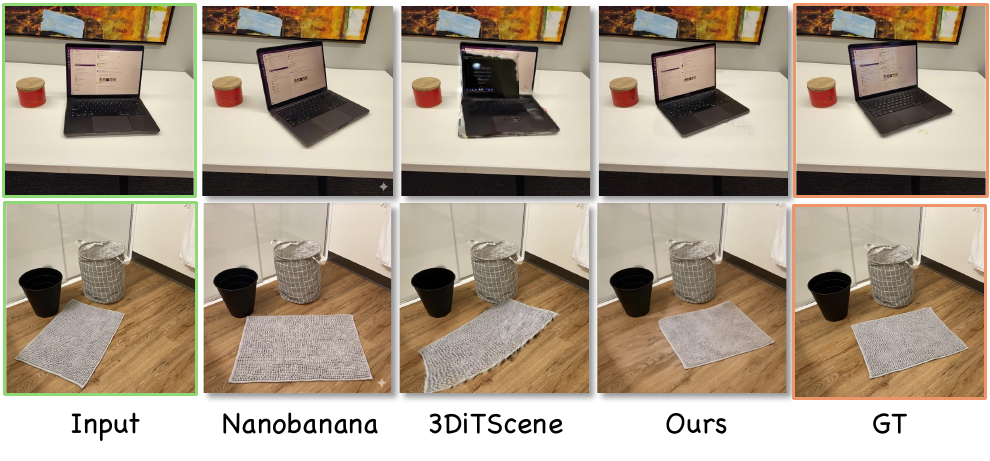}
  \caption{\textbf{Qualitative Comparison on the External 3DEdit-Bench.} Compared to state-of-the-art baselines, GeoEdit effectively generalizes to out-of-domain samples. NanoBanana suffers from geometry and identity loss, while 3DitScene introduces severe ghosting and structural distortion. Our method strictly adheres to the 3D physical constraints and accurately matches the Ground Truth (GT).}
  \label{fig:supp_3dedit_qualitative}
\end{figure}

\vspace{2mm}
\noindent\textbf{Qualitative Comparison on 3DEdit-Bench.} 
To further corroborate our quantitative findings on the external benchmark, we provide visual comparisons sampled from 3DEdit-Bench in Fig.~\ref{fig:supp_3dedit_qualitative}. We specifically compare GeoEdit against the strongest baselines, NanoBanana and 3DitScene, using Ground Truth (GT) as the reference.

As illustrated in Fig.~\ref{fig:supp_3dedit_qualitative}, baseline methods struggle with complex out-of-plane transformations in out-of-domain real-world scenarios. For instance, NanoBanana often fails to maintain the correct geometric perspective or object identity (e.g., the distorted laptop screen). 3DitScene, despite utilizing 3D Gaussian Splatting, frequently produces severe ghosting, geometry collapse, and blending artifacts (e.g., the duplicated laptop and the crumpled rug). In contrast, GeoEdit effectively maintains the rigid physical structure of the target objects and seamlessly harmonizes them within the original background context, achieving results that strictly match the Ground Truth.

\begin{table}[htbp]
  \centering
  \caption{\textbf{Quantitative Comparison on 3DEdit-Bench.} GeoEdit achieves superior performance in both perceptual quality and geometric adherence.}
  \label{tab:3dedit_bench}
  \resizebox{0.95\linewidth}{!}{
  \begin{tabular}{lcccccc}
    \toprule
    Method & PSNR $\uparrow$ & LPIPS $\downarrow$ & DINO $\uparrow$ & CLIP $\uparrow$ & DreamSim $\downarrow$ & Object IoU $\uparrow$ \\
    \midrule
    3DitScene & 20.332 & 0.291 & 0.906 & 0.926 & 0.134 & 0.310 \\
    NanoBanana & 17.512 & 0.304 & 0.945 & \textbf{\textcolor{red}{0.946}} & 0.064 & 0.134 \\
    \textbf{Ours} & \textbf{\textcolor{red}{22.666}} & \textbf{\textcolor{red}{0.207}} & \textbf{\textcolor{red}{0.952}} & 0.912 & \textbf{\textcolor{red}{0.061}} & \textbf{\textcolor{red}{0.460}} \\
    \bottomrule
  \end{tabular}
  }
\end{table}

\vspace{2mm}
\noindent\textbf{Per-Category Results with Confidence Intervals.} 
To provide a granular understanding of our framework's capabilities, Tab.~\ref{tab:per_category} details the performance breakdown across three core spatial transformations: Translation, Rotation, and Camera Move. To ensure rigorous statistical significance as suggested by prior works, we report the 95\% bootstrap confidence intervals (CIs) for key metrics. The results indicate highly consistent generation fidelity across all operation types.

\begin{table}[htbp]
  \centering
  \caption{\textbf{Per-Category Performance on GeoEditBench.} We report the mean metrics along with their 95\% bootstrap confidence intervals (CIs).}
  \label{tab:per_category}
  \resizebox{0.95\linewidth}{!}{
  \begin{tabular}{lccc}
    \toprule
    Category & PSNR (95\% CI) & LPIPS (95\% CI) & Human Pref. (95\% CI) \\
    \midrule
    Translation & 23.8 [22.5, 24.9] & 0.112 [0.08, 0.14] & 4.82 [4.5, 5.0] \\
    Rotation    & \textbf{24.1 [22.8, 25.3]} & \textbf{0.108 [0.08, 0.13]} & \textbf{4.85 [4.6, 5.0]} \\
    Camera Move & 20.7 [18.2, 23.1] & 0.135 [0.10, 0.16] & 4.65 [4.1, 4.9] \\
    \bottomrule
  \end{tabular}
  }
\end{table}

\begin{figure*}[htbp]
  \centering
  \includegraphics[width=\textwidth]{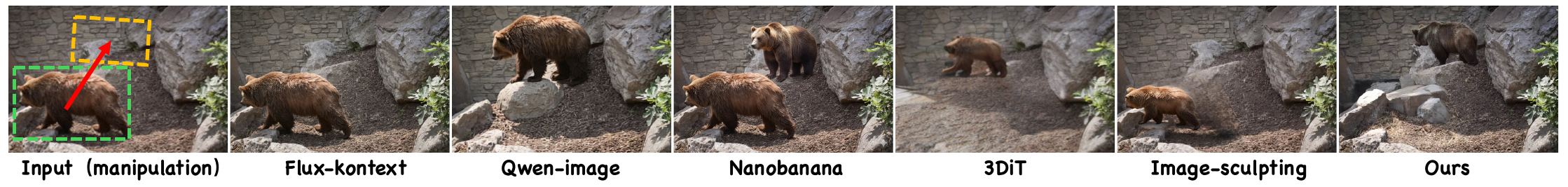} 
  \caption{\textbf{Qualitative comparison on a complex depth-aware translation task.} Given the instruction to translate the target object onto the rock, baseline methods suffer from severe blending artifacts, identity loss, or background geometric distortion. GeoEdit successfully performs the 3D translation while faithfully preserving the original object identity and harmonizing the unedited background context.}
  \label{fig:supp_bear}
\end{figure*}

To further demonstrate the superiority of our approach in handling complex depth-aware manipulations, we present an additional challenging case in Fig.~\ref{fig:supp_bear}. In this scenario, the target object is instructed to be translated across the depth plane onto a specific structural element (\ie, the rock). 

As shown in Fig.~\ref{fig:supp_bear}, existing 2D and 3D-aware methods struggle with this task. Flux-Kontext and Qwen-Image exhibit noticeable blending artifacts and fail to naturally integrate the object into the new depth plane. NanoBanana completely loses the original object identity, hallucinating an entirely different animal instance. 3DiT and Image-Sculpting fail to maintain the structural integrity of the scene, resulting in severe background distortion and a loss of geometric realism. In contrast, our GeoEdit effectively executes the 3D translation, naturally compositing the object onto the rock with accurate shadows while strictly preserving both the object's identity and the unedited background fidelity.

\section{Supplementary Project Page}
\label{sec:demo}

\begin{figure*}[htbp]
  \centering
  \includegraphics[width=0.95\linewidth]{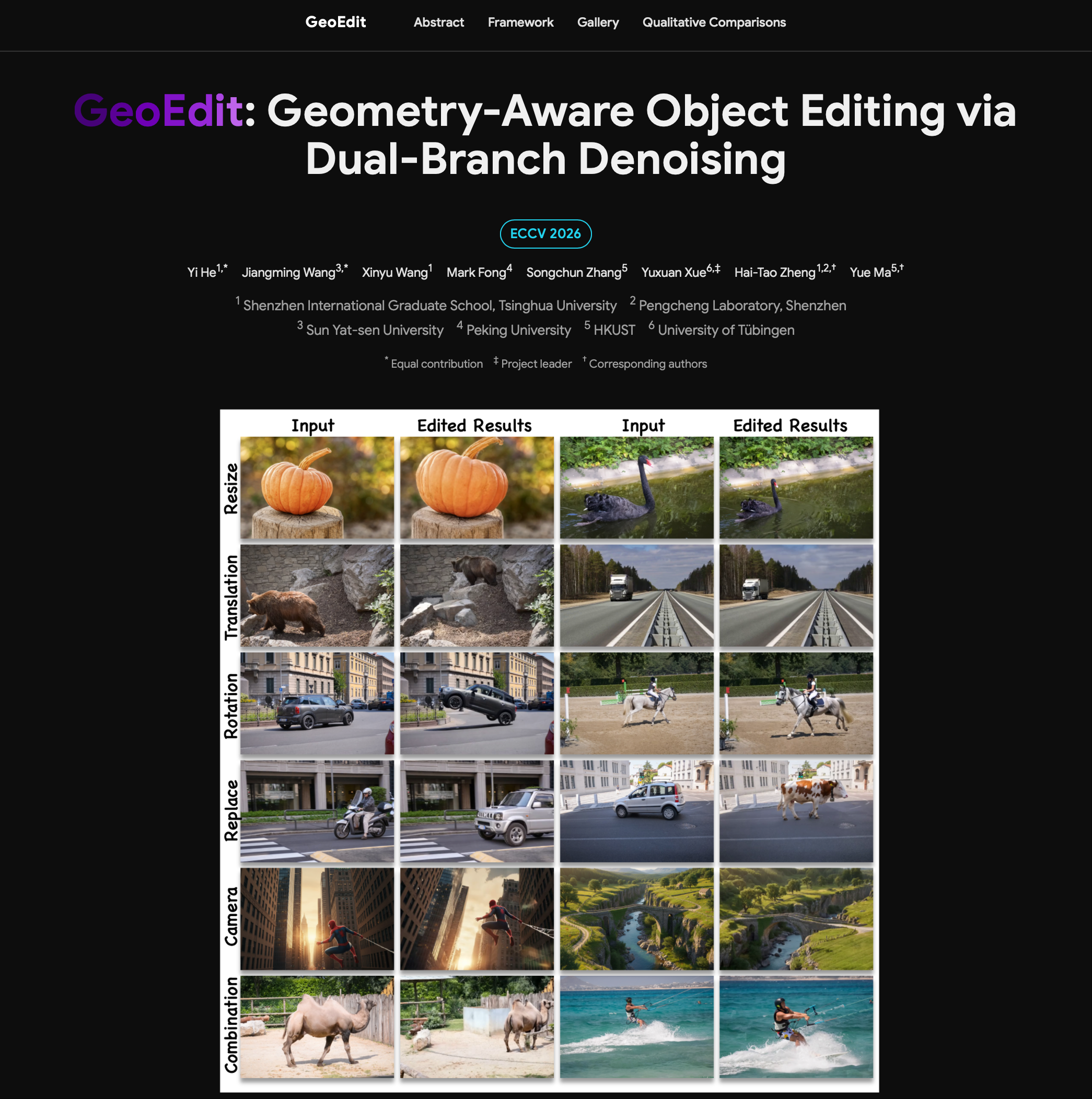} 
  \caption{\textbf{Preview of the Project Page.} Please visit our project page at \url{https://geo-edit.github.io/} to explore an extensive gallery of high-resolution editing results, diverse manipulation scenarios, and comprehensive baseline comparisons.}
  \label{fig:supp_project_page}
\end{figure*}

We provide a project page (\url{https://geo-edit.github.io/}), as previewed in Fig.~\ref{fig:supp_project_page}. This webpage contains an extensive collection of high-resolution qualitative results and additional spatial editing cases across various scenarios. We highly encourage readers to view this supplementary page for a more comprehensive and detailed visual inspection of the generated outputs from our GeoEdit framework and the compared baselines.

\section{Extended Ablation on Frame Count}
\label{sec:frame_ablation}
As discussed in the main text, our framework deliberately extracts the native 81-frame context from the video diffusion backbone to fully exploit its temporal prior for rigid 3D consistency. To validate this design choice, we conducted a visual ablation by forcibly truncating the generation context to 1, 21, 41, and 61 frames. 

\begin{figure*}[htbp]
  \centering
  \includegraphics[width=\linewidth]{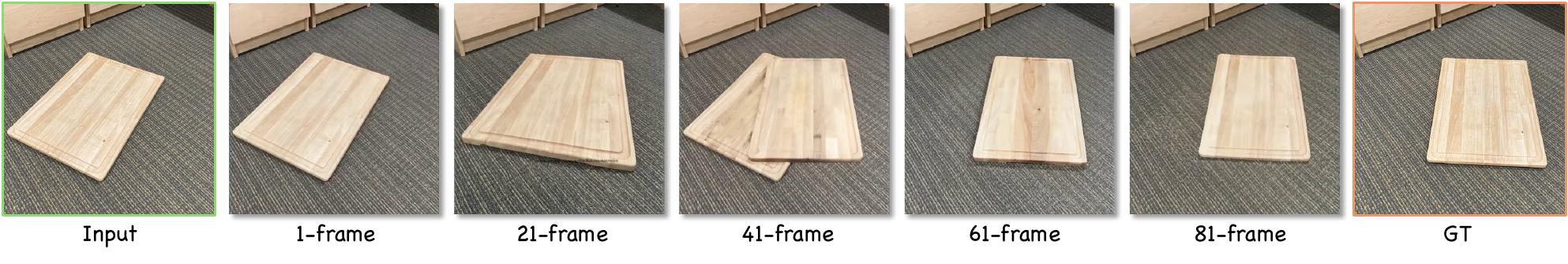}
  \caption{\textbf{Visual Ablation on Frame Count.} Drastically reducing the temporal generation horizon (\eg, 1, 21, or 41 frames) fails to preserve 3D geometric consistency, resulting in severe ghosting and structural deviations from the target proxy. Utilizing the native 81-frame context fully exploits the backbone's temporal prior, strictly locking the geometric rigidity to match the Ground Truth (GT).}
  \label{fig:supp_frame_ablation}
\end{figure*}

As illustrated in Fig.~\ref{fig:supp_frame_ablation}, reducing the frame count severely degrades the structural integrity of the proxy. For instance, generating at 21 or 41 frames leads to catastrophic geometric deviations and visible ghosting effects (\eg, the disjointed ``double-board'' artifacts). While generating fewer frames accelerates inference time, this ablation proves that the full 81-frame context is strictly necessary. It acts as a temporal anchor that safely locks the object's spatial rigidity during the variance-homogeneous injection phase, ultimately producing results that are faithfully aligned with the Ground Truth (GT).

\section{Extended Qualitative Ablations}
\label{sec:extended_ablations}
\begin{figure*}[htbp]
  \centering
  \makebox[\textwidth][c]{\includegraphics[width=1.0\textwidth]{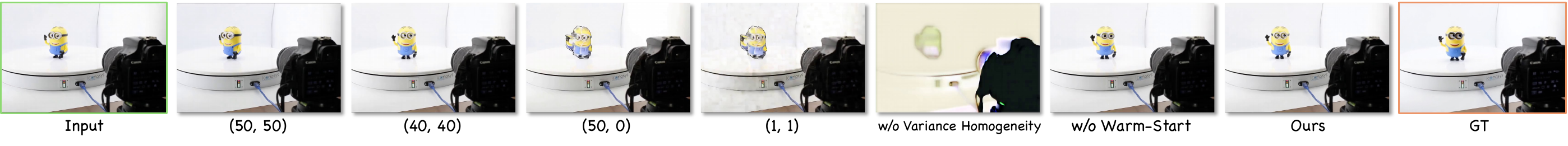}}
  \caption{\textbf{Comprehensive Qualitative Ablation.} We visualize the impact of varying timestep windows and dropping core components. Without variance homogeneity, the generation suffers from catastrophic rendering collapse. Our full method strictly adheres to the 3D manipulation while preserving background photorealism, closely matching the Ground Truth (GT).}
  \label{fig:supp_qualitative_ablation}
\end{figure*}

To supplement the ablation studies in the main text, we provide a comprehensive visual comparison of both timestep configurations and core components on a single complex manipulation task, as shown in Fig.~\ref{fig:supp_qualitative_ablation}. 

The results clearly illustrate the varying failure modes when our symmetric constraints are violated. Relying entirely on the diffusion prior \textbf{(50, 50)} grants excessive generative freedom, resulting in deviations from the intended target geometry. Conversely, excessive proxy injection at low noise levels \textbf{(1, 1)} rigidly forces the 3D shape but completely destroys the background context, yielding severe washing-out effects. 

Regarding the algorithmic components, removing the warm-start initialization \textbf{(w/o Warm-Start)} leads to subtle but noticeable shifts in global illumination and background synthesis. Most critically, substituting our synchronized noise matching with uncalibrated constraints \textbf{(w/o Variance Homogeneity)} severely disrupts the homogeneous latent distribution. As visually evident, this omission causes catastrophic attention leakage and extreme rendering artifacts. In contrast, \textbf{Ours} faithfully resolves these issues, seamlessly harmonizing the background while executing a manipulation that closely matches the \textbf{Ground Truth (GT)}.

\section{Additional Qualitative Results}
\label{sec:cases}
We provide comprehensive visual comparisons across various spatial manipulation scenarios, including complex backgrounds and diverse object categories. As demonstrated in the following figures, our proposed \textbf{GeoEdit} consistently outperforms existing baselines in maintaining geometric constraints, preserving object identity, and harmonizing the unedited background.

\begin{figure*}[htbp]
  \centering
  \includegraphics[width=0.85\textwidth]{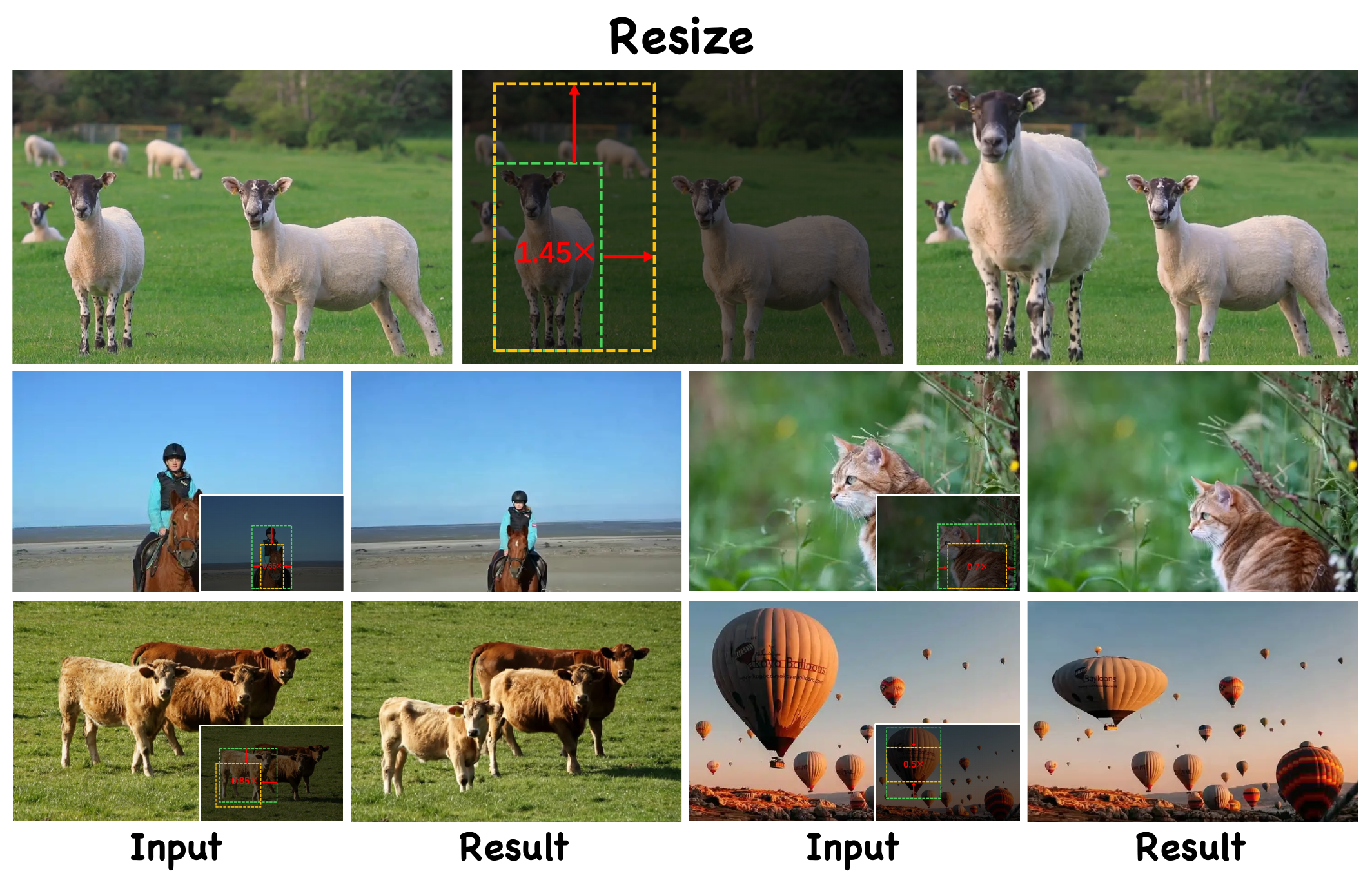}
  \caption{\textbf{Additional Qualitative Results on Object Resizing.} GeoEdit scales the target object while maintaining its geometric proportions and seamlessly inpainting the disoccluded background without artifacts.}
  \label{fig:supp_resize}
\end{figure*}

\begin{figure*}[htbp]
  \centering
  \includegraphics[width=0.85\textwidth]{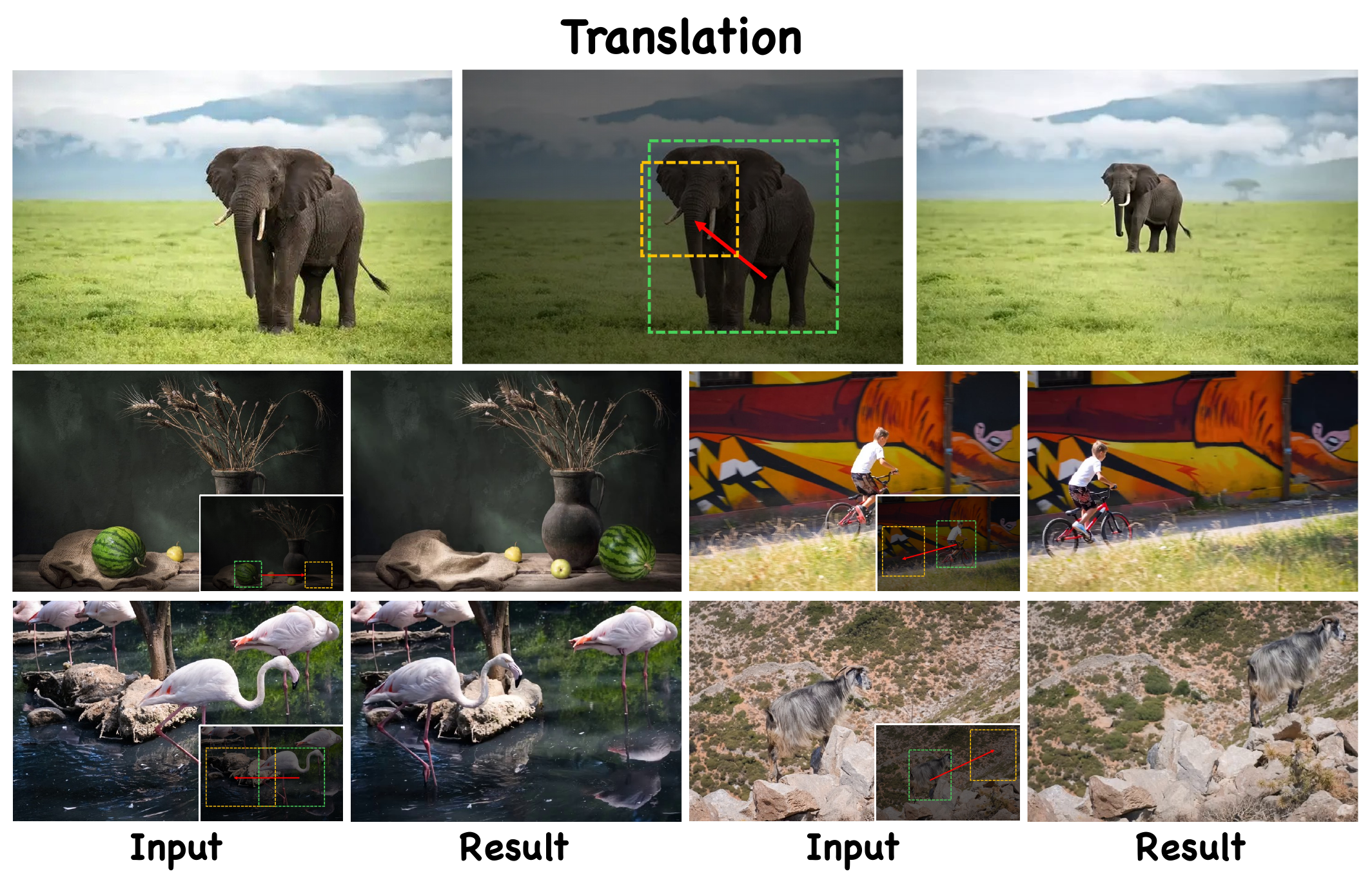}
  \caption{\textbf{Additional Qualitative Results on Object Translation.} GeoEdit seamlessly relocates objects while adhering to strict 3D constraints and cleanly inpainting the original footprints, avoiding the ghosting and perspective errors common in baselines.}
  \label{fig:supp_trans}
\end{figure*}

\begin{figure*}[htbp]
  \centering
  \includegraphics[width=0.85\textwidth]{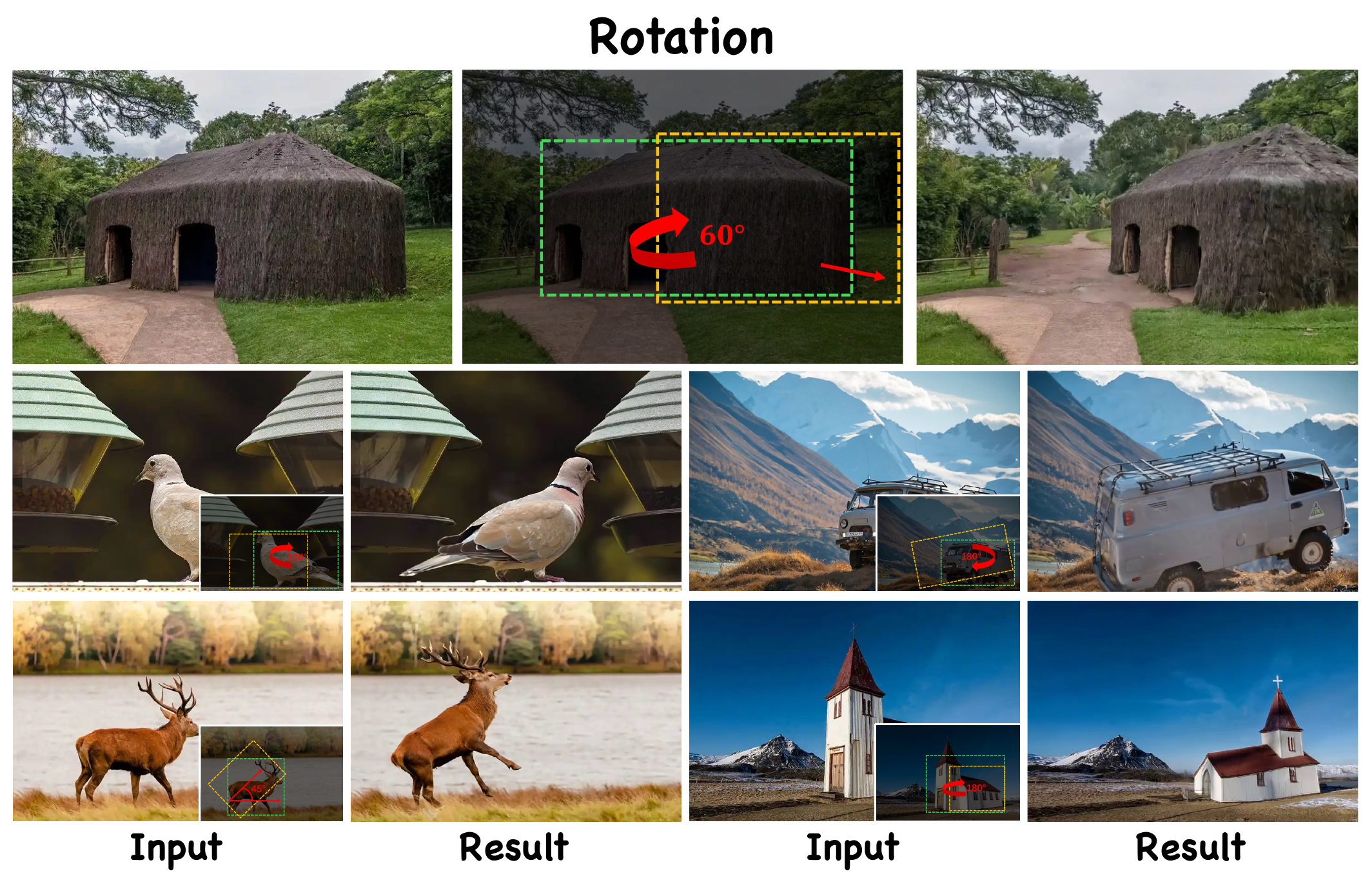}
  \caption{\textbf{Additional Qualitative Results on Object Rotation.} For complex out-of-plane rotations, GeoEdit effectively recovers unseen textures via its multi-view lifting prior, ensuring that the rotated object rigidly adheres to the target 3D orientation without geometric collapse.}
  \label{fig:supp_rotate}
\end{figure*}

\begin{figure*}[htbp]
  \centering
  \includegraphics[width=0.85\textwidth]{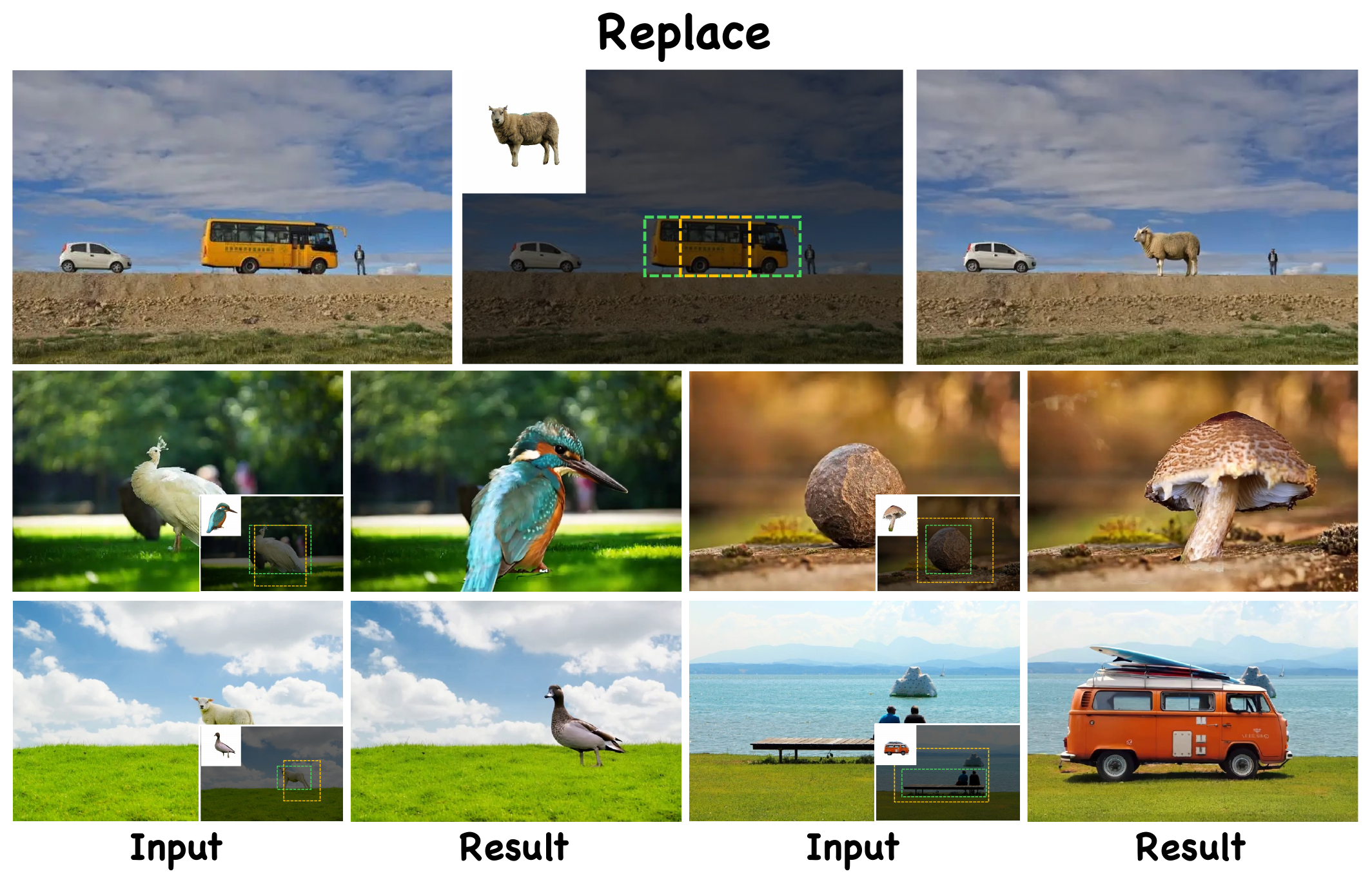}
  \caption{\textbf{Additional Qualitative Results on Object Replacement.} When replacing structural elements, our dual-branch denoising naturally harmonizes the new object into the existing lighting and perspective context of the original scene.}
  \label{fig:supp_replace}
\end{figure*}

\begin{figure*}[htbp]
  \centering
  \includegraphics[width=0.85\textwidth]{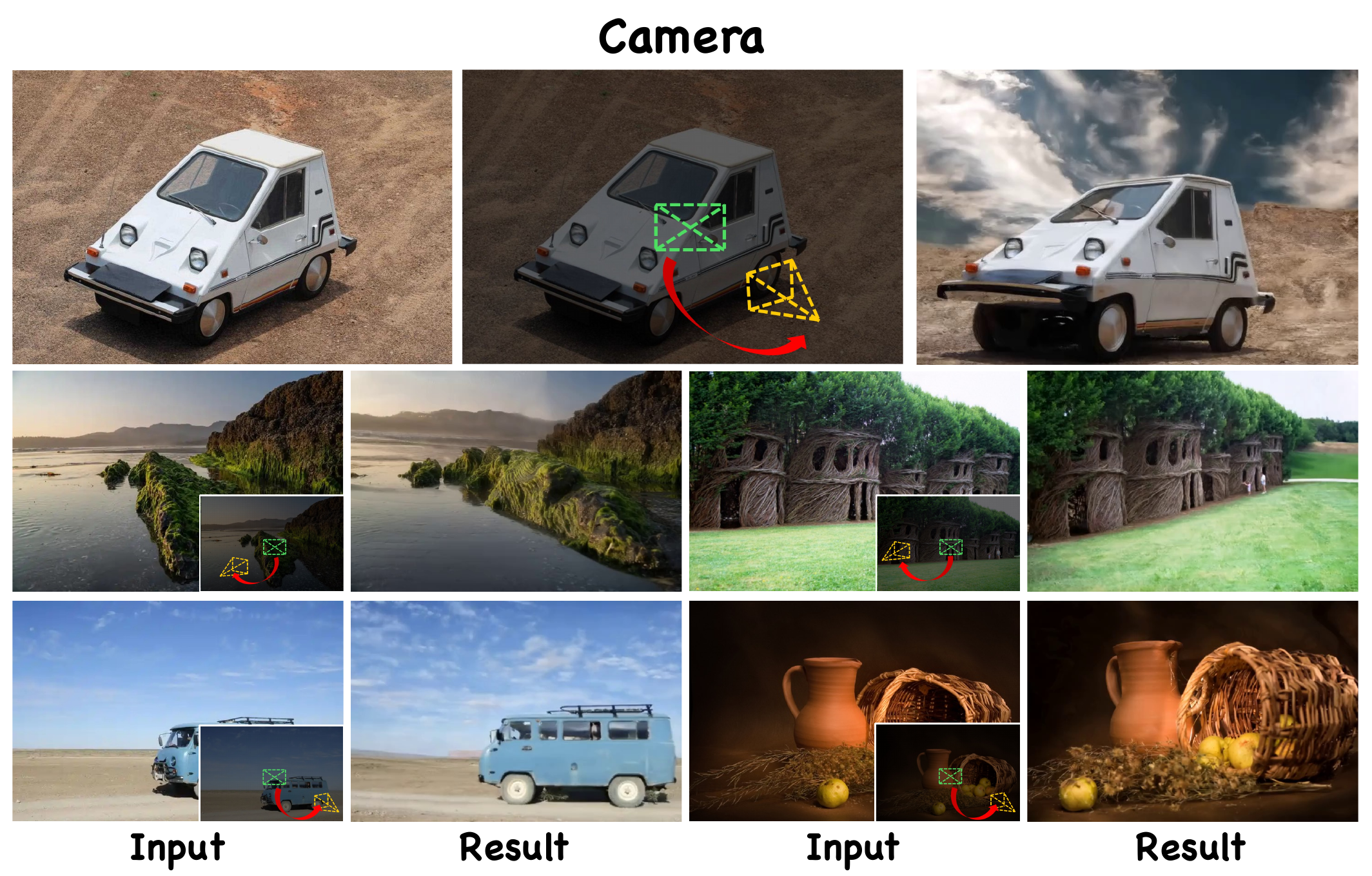}
  \caption{\textbf{Additional Qualitative Results on Camera Viewpoint Variations.} GeoEdit demonstrates robust 3D-aware consistency even when simulating global camera movements, avoiding the spatial distortions common in purely 2D diffusion editing.}
  \label{fig:supp_camera}
\end{figure*}

\begin{figure*}[htbp]
  \centering
  \includegraphics[width=0.85\textwidth]{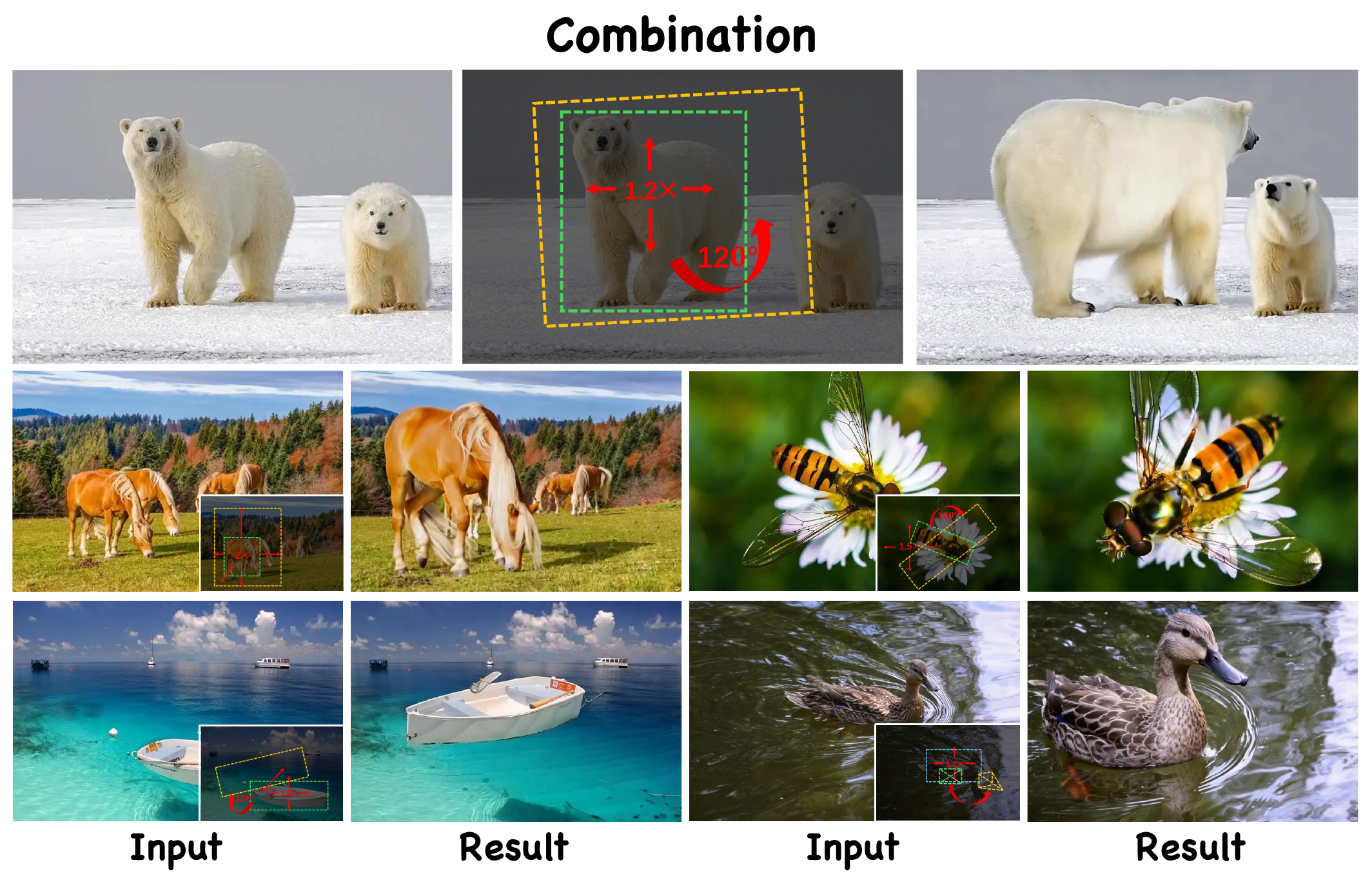}
  \caption{\textbf{Additional Qualitative Results on Combinational Edits.} Executing multiple spatial transformations (\eg, simultaneous rotation and translation) pushes generation to the limit. Our pipeline strictly obeys the combined instructions while preserving photorealism.}
  \label{fig:supp_combination}
\end{figure*}

\section{User Study}
\label{sec:user_study}

Given the inherently subjective nature of generative image editing, we conduct comprehensive qualitative assessments utilizing both human participants and an advanced Vision-Language Model (VLM). To ensure fairness and scientific rigor, both evaluations strictly adhere to the same zero-shot rating protocol and multi-dimensional criteria.

\begin{figure*}[htbp]
  \centering
  \includegraphics[width=0.8\textwidth]{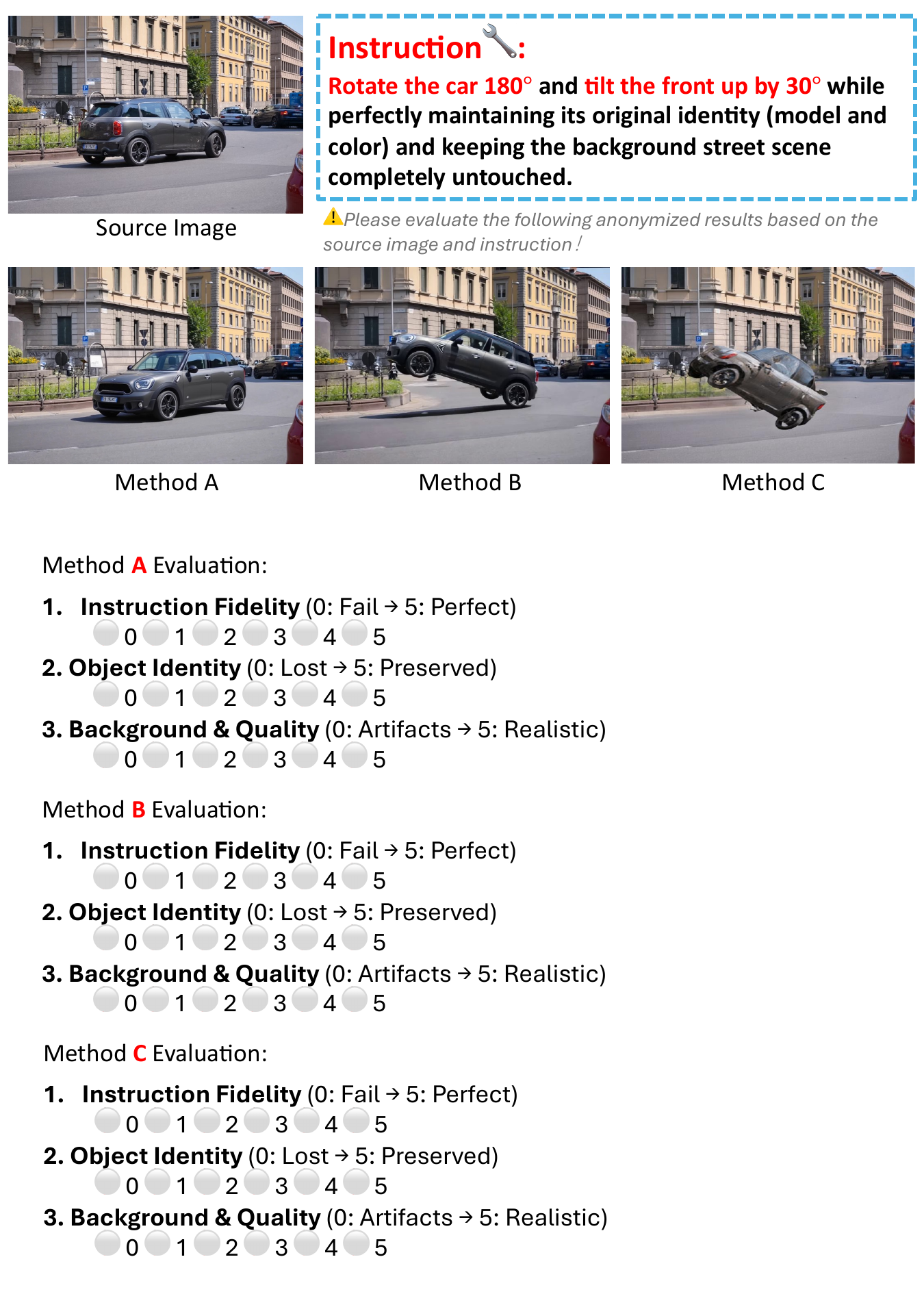} 
  \caption{\textbf{Human Evaluation Interface.} An example of our questionnaire design. For each trial, participants are provided with the source image and a strict manipulation instruction. To prevent fatigue and bias, 3 randomly sampled methods (out of 6) are presented as Method A, B, and C. Participants evaluate the anonymized results across three specific criteria on a 0-5 scale.}
  \label{fig:user_study_ui}
\end{figure*}

\vspace{1mm}
\noindent\textbf{1. Human Evaluation Protocol.} \\
We recruited 25 independent volunteers for a rigorous, single-blind Mean Opinion Score (MOS) study. To prevent visual fatigue and ensure high-quality feedback (reducing cognitive load), we designed a streamlined sampling strategy:
\begin{itemize}
    \item \textbf{Task Assignment:} Each participant was randomly assigned 3 distinct spatial editing cases (image-instruction pairs) covering diverse transformations.
    \item \textbf{Randomized Sub-sampling:} For each case, instead of overwhelmingly presenting all 6 methods at once, we randomly sampled 3 generated results from the full pool (our GeoEdit and 5 baselines). 
    \item \textbf{Anonymization:} As illustrated in Fig.~\ref{fig:user_study_ui}, the selected results were displayed side-by-side and strictly anonymized as \textit{``Method A''}, \textit{``Method B''}, and \textit{``Method C''}. Participants could only evaluate the outputs based on their visual alignment with the provided source image and the strict textual instruction.
\end{itemize}

\vspace{1mm}
\noindent\textbf{2. VLM Expert Evaluation Protocol.} \\
To complement the human study with a scalable and unbiased metric, we employed a Gemini-family multimodal model~\cite{gemini} as an expert VLM judge. The AI evaluation faithfully mirrored the human protocol:
\begin{itemize}
    \item \textbf{System Prompting:} The VLM was explicitly prompted to act as an ``impartial and professional image quality scorer.'' It was provided with the exact same triplet of inputs: the original source image, the precise manipulation instruction, and the generated output.
    \item \textbf{Scoring Execution:} The VLM analyzed visual coherence, geometric consistency, and instruction alignment, outputting a discrete score without any contextual bias regarding the model identities.
\end{itemize}

\vspace{1mm}
\noindent\textbf{3. Evaluation Metrics (0-5 Likert Scale).} \\
Both human volunteers and the AI judge were required to rate the generated results on a discrete scale from 0 (Completely Fail) to 5 (Perfect) across three finely disentangled dimensions:
\begin{itemize}
    \item \textbf{Instruction Fidelity:} Evaluates how accurately the image reflects the given spatial manipulation command.
    \item \textbf{Object Identity Preservation:} Assesses whether the manipulated object faithfully retains its original semantic attributes, fine-grained textures, and instance-level identity.
    \item \textbf{Background \& Image Quality:} Rates the overall photorealism, checking whether the disoccluded regions are seamlessly inpainted and free from ghosting artifacts or perspective distortions.
\end{itemize}

\vspace{1mm}
\noindent\textbf{Aggregation and Quality Control.} \\
For human ratings, we implemented strict quality control by monitoring completion time. Responses falling significantly below a reasonable time threshold were discarded. Extreme statistical outliers for specific image pairs were also removed before computing the averages. The valid scores from both human evaluators and the VLM judge were aggregated respectively to compute the final Mean Preference Scores. As reported in the main text, our GeoEdit secures the highest preference scores across both human and AI evaluations, corroborating its robust perceptual superiority.

\section{Failure Cases Analysis}
\label{sec:failure_cases}
\begin{figure*}[htbp]
  \centering
  \includegraphics[width=0.8\linewidth]{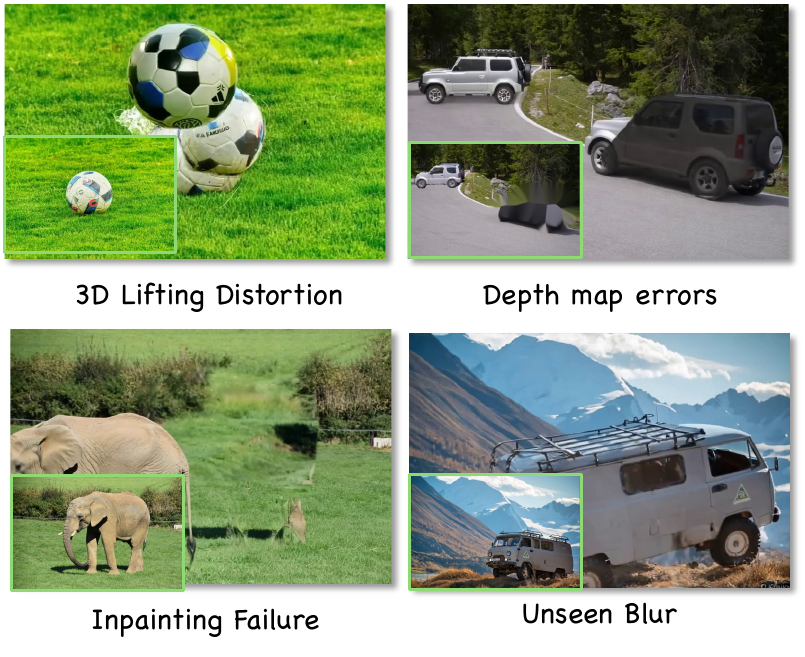} 
  \caption{\textbf{Typical Failure Cases.} (Top) 3D lifting distortion causes the geometry proxy to collapse. (Bottom) Extreme spatial edits exceed the background inpainting capacity, resulting in semantic artifacts.}
  \label{fig:supp_failure}
\end{figure*}

While GeoEdit excels in most spatial manipulations, we provide visual examples of typical failure modes in Fig.~\ref{fig:supp_failure}. These failures generally stem from two bottlenecks:
\begin{enumerate}
    \item \textbf{Geometry Extraction Flaws:} When the monocular depth estimator or the multi-view lifting model produces distorted structures, the resulting 3D proxy becomes warped, leading to downstream alignment errors (see Fig.~\ref{fig:supp_failure} top).
    \item \textbf{Extreme Spatial Edits:} Executing massive translations or extreme rotations can expose vast disoccluded regions or fully unseen back-facets. If these regions exceed the generative prior's hallucinatory capacity, it can result in blurred textures or inpainting artifacts (see Fig.~\ref{fig:supp_failure} bottom).
\end{enumerate}

\section{Social Potential Impact and Limitations}
\label{sec:impact_and_limitations}

\paragraph{Social Impact.} 
Our framework democratizes complex 3D-aware image editing, significantly lowering the technical barrier for creators in fields such as augmented/virtual reality (AR/VR), digital advertising, and immersive design. By eliminating the need for professional 3D modeling expertise, GeoEdit empowers users to intuitively manipulate visual content. However, as with other high-fidelity generative models, the ability to seamlessly alter object states and scene geometry raises concerns regarding visual misinformation and malicious content spoofing. To mitigate these dual-use risks, we advocate for the integration of our framework with robust digital provenance protocols and proactive digital watermarking techniques to ensure responsible deployment and content authenticity.

\paragraph{Limitations.} 
While GeoEdit achieves robust geometric alignment and high-fidelity denoising, it inherently inherits certain limitations from its foundational components. First, due to the cascaded nature of our pipeline, the final generation quality is heavily bottlenecked by the accuracy of the intermediate 3D point cloud reconstruction. Inaccuracies in monocular depth estimation or novel-view synthesis can introduce structural distortions that inevitably propagate to the final composite. Consequently, handling extreme occlusions---such as rotating an object behind dense foliage---remains challenging. Second, the reliance on a large-scale video diffusion backbone to enforce strict 3D rigid priors incurs a noticeably higher computational overhead and inference latency compared to single-step 2D image editors. Finally, because our dual-branch denoising relies on a learned generative prior rather than a deterministic physical rendering engine, strictly simulating view-dependent lighting effects (\eg, accurate reflections on mirrors or glass) is currently beyond the scope of this work. Future research will focus on developing more lightweight, end-to-end architectures and integrating physically based rendering (PBR) guidance to resolve these challenges.

\end{document}